\documentclass[sigconf]{acmart}
\usepackage{amsmath}
\usepackage{bm}
\usepackage{mathrsfs}
\usepackage{amsthm}
\usepackage{amsfonts}
\usepackage{subfigure}
\usepackage{multirow}
\usepackage{url}
\usepackage{array}
\usepackage{colortbl}
\usepackage[ruled,linesnumbered,vlined]{algorithm2e}
\usepackage{algorithmic}
\usepackage{booktabs}
\usepackage{longtable}
\usepackage{multirow}
\usepackage{multicol}

\AtBeginDocument{%
  \providecommand\BibTeX{{%
    \normalfont B\kern-0.5em{\scshape i\kern-0.25em b}\kern-0.8em\TeX}}}
\setcopyright{acmcopyright}
\copyrightyear{2024} 
\acmYear{2024} 
\setcopyright{acmlicensed}\acmConference[WSDM '24]{Proceedings of the Seventeenth ACM International Conference on Web Search and Data Mining}{March 4-8, 2024}{Merida, Mexico}
\acmBooktitle{Proceedings of the Seventeenth ACM International Conference on Web Search and Data Mining (WSDM '24), March 4-8, 2024, Merida, Mexico}
\acmPrice{15.00}
\acmDOI{10.1145/xxxxxxx.xxxxxxx}
\acmISBN{978-1-4503-9407-x/xx/xx}

\newcommand{\framework}{MultiSPANS}

\begin{document}
% \title{\framework{}: Multi-Range Spatial-Temporal Transformer Network for Traffic Forecast }
\title{\framework{}: A Multi-range Spatial-Temporal Transformer\\ Network for Traffic Forecast via Structural Entropy Optimization}

% \author{Anonymous}
% \affiliation{%
%   \institution{Beihang University}
%   \city{Beijing}
%   \country{China}
% }

\author{Dongcheng Zou}
\affiliation{%
  \institution{Beihang University}
  \city{Beijing}
  \country{China}
}
\email{zoudongcheng@buaa.edu.cn}

\author{Senzhang Wang}
\affiliation{%
  \institution{Central South University}
  \city{Changsha, Hunan}
  \country{China}
}
\email{szwang@csu.edu.cn}
\authornote{Corresponding authors}

\author{Xuefeng Li}
\affiliation{%
  \institution{Beihang University}
  \city{Beijing}
  \country{China}
}
\email{xuefengli@buaa.edu.cn}

\author{Hao Peng}
\authornotemark[1]
\affiliation{%
  \institution{Beihang University}
  \city{Beijing}
  \country{China}
}
\email{penghao@buaa.edu.cn}

\author{Yuandong Wang}
\authornotemark[1]
\affiliation{%
  \institution{Tsinghua University}
  \city{Beijing}
  \country{China}
}
\email{wangyd2021@tsinghua.edu.cn }

\author{Chunyang Liu}
\affiliation{%
  \institution{Didi chuxing}
  \city{Beijing}
  \country{China}
}
\email{liuchunyang@didiglobal.com}

\author{Kehua Sheng}
\affiliation{%
  \institution{Didi chuxing}
  \city{Beijing}
  \country{China}
}
\email{shengkehua@didiglobal.com}

\author{Bo Zhang}
\affiliation{%
  \institution{Didi chuxing}
  \city{Beijing}
  \country{China}
}
\email{zhangbo@didiglobal.com}

\renewcommand{\shortauthors}{D. Zou et al.}

\theoremstyle{definition}
\newtheorem{define}{Definition}[]

\begin{abstract}
Traffic forecasting is a complex multivariate time-series regression task of paramount importance for traffic management and planning. 
% However, existing approaches often struggle to model multi-range spatiality and temporality at both global and local scales.
However, existing approaches often struggle to model complex multi-range dependencies using local spatiotemporal features and road network hierarchical knowledge.
To address this, we propose \framework{}. 
First, considering that an individual recording point cannot reflect critical spatiotemporal local patterns, we design multi-filter convolution modules for generating informative \textit{ST-token} embeddings to facilitate attention computation.
Then, based on \textit{ST-token} and spatial-temporal position encoding, we employ the Transformers to capture long-range temporal and spatial dependencies. 
Furthermore, we introduce structural entropy theory to optimize the spatial attention mechanism.
Specifically, The structural entropy minimization algorithm is used to generate optimal road network hierarchies, i.e., encoding trees. 
Based on this, we propose a relative structural entropy-based position encoding and a multi-head attention masking scheme based on multi-layer encoding trees.
% To address this, we propose \framework{}, a novel framework that utilizes a multi-filter convolution module to extract short-range local patterns and spatial-temporal Transformers to capture long-range global dependencies while innovatively introducing structural entropy optimization into multi-range spatial attention.
% Specifically, we use multi-size temporal and multi-hop graph convolution filters to generate informative local spatiotemporal representations with extensive channels. 
% Then we employ the Transformers with multi-headed self-attention to adaptively capture long-range dependencies separately in the temporal and spatial dimensions.
% To better preserve prior road network topology information, we employ a structural entropy-guided graph structure perception mechanism to generate the hierarchical abstraction and obtain a hierarchical relative position encoding and multi-level attention masks.
Extensive experiments demonstrate the superiority of the presented framework over several state-of-the-art methods in real-world traffic datasets, and the longer historical windows are effectively utilized.
The code is available at 
\href{https://github.com/SELGroup/MultiSPANS}{https://github.com/SELGroup/MultiSPANS}.
\end{abstract}



\maketitle

\section{INTRODUCTION}\label{sec:intro}

\begin{figure}[t]
    \centering
    \includegraphics[width=0.46\textwidth]{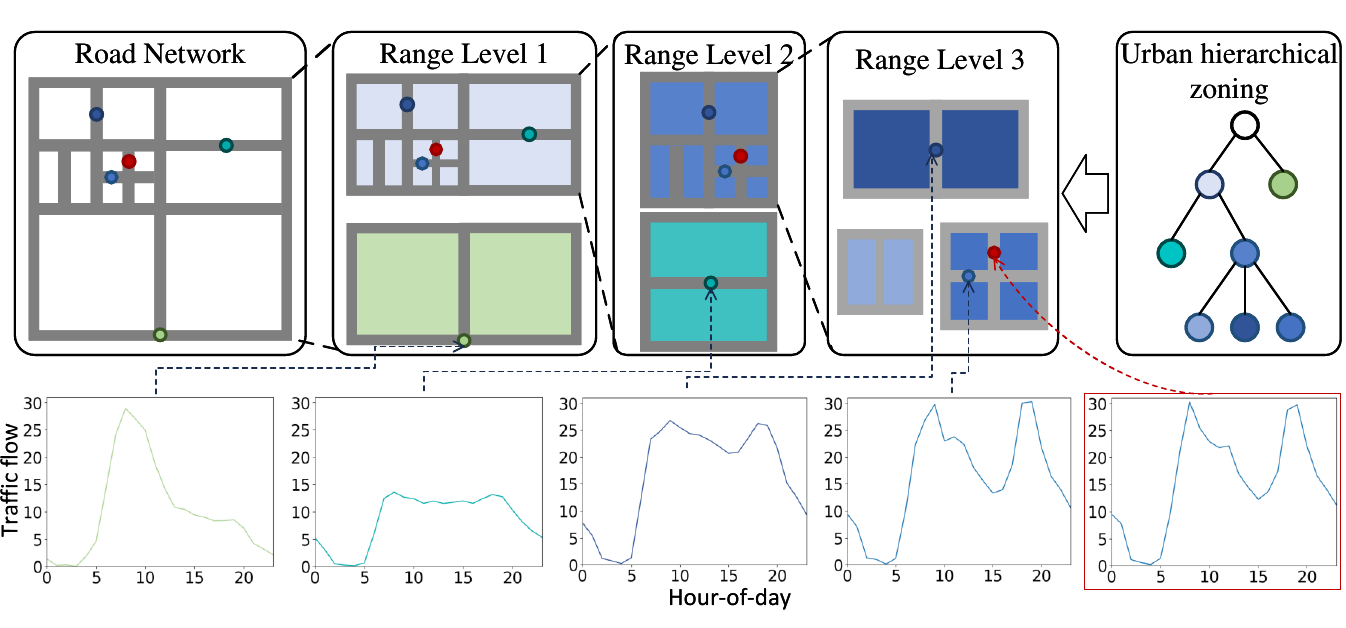}
    \caption{An toy example of the hierarchical urban zoning and its impact on traffic flow.}
    \label{fig:sample}
\end{figure}

Transportation is a complex real-world system that includes people, vehicles, road network sensors, and other components, with a wealth of temporal and spatial connections.
As urbanization continues to advance, there is an increasing demand for more precise analysis of traffic data to improve the efficiency of transportation systems.
To address the growing complexity of traffic-related tasks, deep learning approaches have been widely employed for route planning~\cite{luo2021stan,sun2020go,lian2020geography}, flow prediction~\cite{liDiffusionConvolutionalRecurrent2022,guoAttentionBasedSpatialTemporal2019,zhengGMAN2020,jiangPDFormer2023}, accident prediction~\cite{yuan2018hetero,wang2021gsnet,huang2019deep}, vehicle scheduling~\cite{wang2019originmatrix,ye2021coupled}, etc.
One of the fundamental technologies for intelligent transportation is traffic state forecast, which can be considered as a multivariate time series regression task. 
It involves modeling temporal and spatial dependencies to predict future traffic situations (e.g., flow, speed, or occupancy) based on prior road networks, historical observations, and external traffic-related information.

%%% 目前主流交通预测时序建模方法包括（RNN，TCN），基于路网的空间建模方法主要是GNN，
Current fundamental time-series methods for traffic forecast tasks include Recurrent Neural Networks (RNNs)~\cite{liDiffusionConvolutionalRecurrent2022,zhaoTGCNTemporalGraph2020, BaiAdaptive2020} and Temporal Convolutional Networks (TCNs)~\cite{wuConnectingDotsMultivariate2020,wuGraphWaveNetDeep2019, LuSpatiotemporalAdaptiveGated2020}, while Graph Neural Networks (GNNs) are commonly used to factor the spatial attributes~\cite{liDiffusionConvolutionalRecurrent2022,zhaoTGCNTemporalGraph2020,wuGraphWaveNetDeep2019}.
These works still face challenges, such as difficulty in modeling long-range dependencies ~\cite{abu2019mixhop,chen2020simple}, dealing with time-varying graphs~\cite{guoLearningDynamicsHeterogeneity2022}, and coping with unreliable structures~\cite{wuConnectingDotsMultivariate2020}.
Recently, Transformer~\cite{VaswaniAttention2017} has been widely used in spatiotemporal tasks to address existing issues.
However, these frontier Transformer-based methods have two problems corresponding to time-series and graph learning. 
First, Transformers may not be as effective as expected in handling long time-series data~\cite{zengAreTransformersEffective2022,liMTSMixersMultivariateTime2023}. 
It is possibly because the information in discrete time points is insufficient to learn pairwise attention and model higher-order global temporality ~\cite{nieTimeSeriesWorth2023,jiangPDFormer2023}.
Second, Transformers have difficulty in directly utilizing the graph structure. 
Mainstream approaches include fusing GNN and Transformer output~\cite{zhengGMAN2020,xuSpatialTemporalTransformerNetworks2021} or obtaining simple attention masks/encoding ~\cite{jiangPDFormer2023,ying2021transformers} from networks.
These structure learning mechanisms for Transformers are designed without theoretical guidance and may ignore the rich structural information.

To address both issues in spatial-temporal Transformers, we aim to improve the network to capture rich spatiotemporal dependencies from multiple ranges.
Motivated by patching techniques in the visual~\cite{DosovitskiyViT2021} Transformer, we aim to extract and aggregate multi-frequency local spatiotemporal signals to obtain more representational \textit{ST-tokens} as the basis for effective attention computation. 
Further, we expect Transformer to focus on urban functional zoning impact on traffic state (i.e., greater correlation in the same section). 
As shown in Fig. ~\ref{fig:sample}, functional zoning is naturally hierarchical, reflected in the road network, hard to predefine, and highly correlated with traffic states, e.g., roads in the same high-level community have similar flow characteristics.
Therefore, we introduce the structural entropy theory to measure the uncertainty of the road network and obtain the hierarchical zoning unsupervisedly and adaptively.
Specifically, we propose ~\framework{}: a \textbf{Multi}-range \textbf{S}patiotemporal \textbf{P}rediction \textbf{A}ttention \textbf{N}etwork with \textbf{S}tructural entropy optimization. 
First, we design a lightweight multi-filter convolution module comprising temporal filters graph filters for \textit{ST-tokens} with extensive local information.
Then, we organize the network by interleaving multiple temporal and spatial Transformers to enhance the model's fitting capability toward complex traffic data.
Moreover, an innovative hierarchical graph perception mechanism based on structural entropy is presented.
Structural entropy~\cite{li2016structural} can measure the complexity of a road network and guide the optimal graph hierarchical abstraction by creating the encoding tree~\cite{li2018decoding}. 
According to the structural entropy and multi-level encoding tree, we devised hierarchical correlation scores to identify the nodes' position in the hierarchical community, and multi-level attention masks to learn the relevance at different structural levels separately on each attention head.
The main contributions are outlined below:
\begin{itemize}
	\item A novel and effective spatial-temporal Transformer network, \framework{}, is proposed for a more accurate and versatile traffic state forecast, which addresses current issues. Experiments validate that our method achieves new SOTA in real-world road network datasets.
	\item A practical and pluggable spatial-temporal convolutional module is proposed to obtain informative \textit{ST-tokens} for Transformers in spatiotemporal tasks. It can embed longer historical windows with high computational efficiency to enhance the model's ability to handle long time series.
	\item The structural entropy theory is first exploited to optimize the spatial attention mechanism, which mines the hierarchical structure of the road networks. Visualization study shows that our method can intuitively model multi-range spatial dependencies and discover more relative patterns.
\end{itemize}

\begin{figure*}[t]
    \centering
    \includegraphics[width=1.02\textwidth]{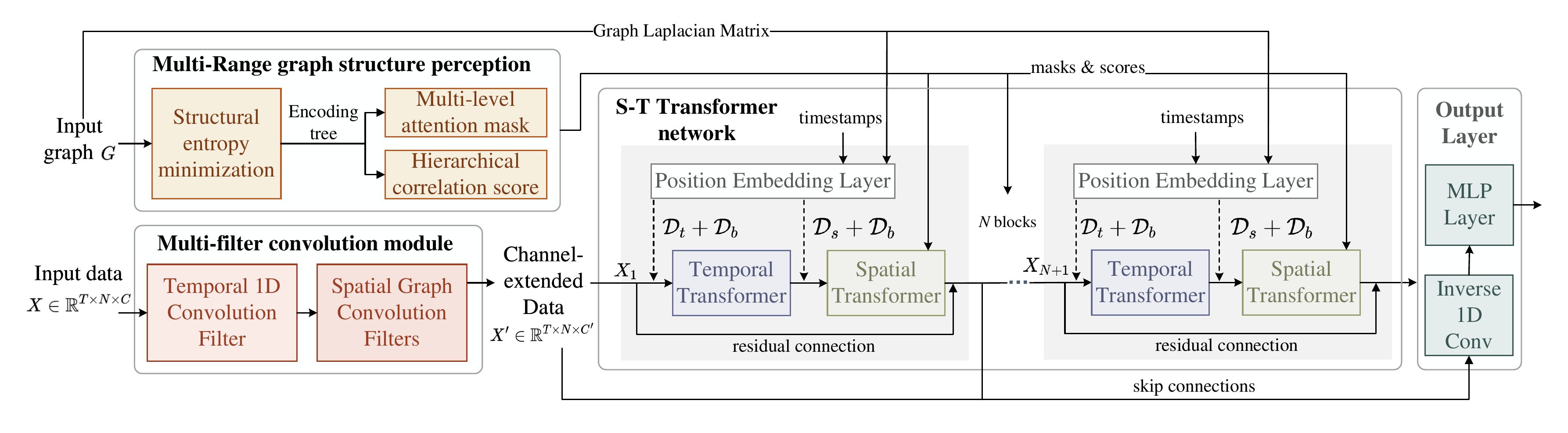}
    \caption{The overall architecture of ~\framework{}.}
    \label{fig:framework}
\end{figure*}

\section{PRELIMINARIES}
% In this section, we will provide relevant definitions and formulate the problem of traffic state forecasting.
% This section will provide notations and definitions of the traffic state data, traffic forecast problem, graph, and structural entropy. 

% \begin{define}
% \textbf{Traffic State Data and Traffic State Forecasting}
% \subsection{Traffic State Data and Traffic Forecast}
\subsection{Problem Definition}
The $C$-channel (speed, flow, occupancy, etc.) traffic state signal collected by the $n$-th sensor at the moment $t$ (i.e., atomic data point) can be represented by the vector $x_{n,t} \in \mathbb{R}^C$. 
The traffic state feature in a time window of width $T$ (starting from moment $t$) for a road network with $N$ sensor nodes can be represented as:
\begin{equation}
    X_{[t,t+T]} = 
    \begin{bmatrix}
        \begin{bmatrix}
            x_{1,t+1} \\ x_{2,t+1} \\ \cdots \\ x_{N,t+1}
        \end{bmatrix}
        & 
        \begin{bmatrix}
            x_{1,t+2} \\ x_{2,t+2} \\ \cdots \\ x_{N,t+2}
        \end{bmatrix}
        &
        \cdots
        &
        \begin{bmatrix}
            x_{1,t+T} \\ x_{2,t+T} \\ \cdots \\ x_{N,t+T}
        \end{bmatrix}
    \end{bmatrix}
    \in \mathbb{R}^{ T \times N \times C}.
\end{equation}
% \begin{equation}
%     X_{[t,t+T]} = [
%             [x_{1,t+1},  \dots,  x_{N,t+1}],
%         \dots,
%             [x_{1,t+T}  ,\dots,  x_{N,t+T}]
%            ]
%     \in \mathbb{R}^{ T \times N \times C}.
% \end{equation}
The traffic state forecasting problem aims to predict future traffic states according to historical observations, prior structure, and additional information, which can be formalized as:
\begin{equation}
    \hat{X}_{[t,t+T']} = f_{\theta} \left ( X_{[t-T,t]},  A_{t-T}, G \right ),
\end{equation}
where $f_{\theta}$ is model with parameter $\theta$, $\hat{X}_{[t,t+T']}$ is the predicted time window of width $T'$, and $A_{t-T}$ is the addition information of the historical window. $G$ denotes the topology structure, which can be road network maps or dynamic graph sequences.

\subsection{Graphs and Structural Entropy}
Let $G = \{V, E\}$ denote a graph, where $V$ is the set of $N$ vertices ~\footnote{Vertices are defined in the graph and nodes are in the tree.} and $E \subseteq V \times V$ is the edge set. $\mathrm {A} \in \mathbb{R}^{N \times N}$ denotes the adjacency matrix of $G$, where $\mathrm {A}_{ij}$ is referred to as the weight of the edge from vertex $i$ to vertex $j$. The degree of vertex $v_i \in V$ is defined as $d(v_i) = \sum_{j}\mathrm {A}_{ij}$, and $D = \mathrm {diag}(d(v_1),d(v_2),\dots,d(v_N))$ refers to the degree matrix.
Recent research by Li and Pan ~\cite{li2016structural} has systematically presented the structural information theory, aiming to measure the uncertainty and information embedded in graphs and obtain the informative hierarchical structures for graph compression.
The theory mainly consists of two parts: Encoding Tree and Structural Entropy.

\noindent \textbf{Encoding Tree}
An encoding tree is a hierarchy that encodes and compresses graphs. For the graph $G=\{V,E\}$, the encoding tree $\mathrm{T}$ rooted at node $\mathrm{\lambda}$ is defined with the following properties: 
1) For each node $\mathrm{\alpha}$ in $\mathrm{T}$, its associated vertex (e.g., the physical node in graph $G$) set is defined as $\mathcal{T}_\alpha \subseteq V$. 
2) For each node $\mathrm{\alpha}$, its parent node is denoted as $\mathrm{\alpha}^-$ and its $i$-th children node is denoted as $\mathrm{\alpha}^{\left \langle i \right \rangle }$ ordered from left to right as $i$ increases.
3) For each non-leaf node $\mathrm{\alpha}$ with $N$ children, all vertex subset $\mathcal{T}_{\mathrm{\alpha}^{\left \langle i \right \rangle}}$ satisfy $\mathcal{T}_\mathrm{\alpha} = {\textstyle \bigcup_{i=1}^{N}} \mathcal{T}_{\mathrm{\alpha}^{\left \langle i \right \rangle}}$ and ${\textstyle \bigcap_{i=1}^{N}} \mathcal{T}_{\mathrm{\alpha}^{\left \langle i \right \rangle}} = \varnothing$.
Thus, the encoding tree abstracts and encodes the graph into a hierarchical community structure.

\noindent \textbf{Structural Entropy}
Structural entropy is determined by the encoding tree and the graph together, which can be formulated as follows:
\begin{equation}\label{eq:HT}
% \vspace{-0.1em}
H^{\mathrm{T}}(G) = \sum_{\mathrm{\alpha} \in \mathrm{T},\mathrm{\alpha} \ne \mathrm{\lambda}} {H^{\mathrm{T}}(G;\mathrm{\alpha})} = -\sum_{\mathrm{\alpha} \in {\mathrm{T}},\mathrm{\alpha} \ne \lambda} {\frac{g_\mathrm{\alpha}}{vol(G)}\log_{2}{\frac{\mathcal{V}_{\mathrm{\alpha}}}{\mathcal{V}_{\mathrm{\alpha}^-}}}},
\end{equation}
where $g_\mathrm{\alpha}$ is the sum weights of edges from the vertices outside $\mathcal{T}_\mathrm{\alpha}$ to those inside $\mathcal{T}_\mathrm{\alpha}$. 
$vol(G)$ is the sum degree of all vertices in $G$, and $\mathcal{V}_\mathrm{\alpha}$ is the sum degree in $\mathcal{T}_\mathrm{\alpha}$. 
The encoding tree that minimizes the graph's structural entropy compresses the most knowledge. Therefore, taking the total information in the graph as constant, it is optimal to represent the essential graph hierarchical structure.
\section{Proposed Method}
% This section presents the overall architecture of ~\framework{} before elaborating on specific designs that facilitate the global and local spatiotemporal dependencies modeling, including the multi-filter convolution module and the spatial-temporal Transformer encoders with the structural entropy-guided hierarchical graph structure perception mechanism for better spatial attention. 

% elaborating on how our approach can efficiently use global and local dependencies through the multi-filter convolution module and the structural entropy guide attention module.

\subsection{Overall Architecture}
\label{sec:overall}

Fig.~\ref{fig:framework} depicts the comprehensive architecture, encompassing three primary sub-modules: the multi-filter convolutional (MFCL) module, the spatial-temporal (ST) Transformers, and the hierarchical graph perception mechanism.
Firstly, we employ the MFCL module to obtain \textit{ST-tokens}, including multiple 1D filters to enhance temporal signals at diverse frequencies and multi-hop graph convolutional filter to aggregate neighborhood signals (\S~\ref{sec:MFSTC}). 
Next, we model complex dependencies with the Transformer network, consisting of a stack of ST encoders with residual connections. 
Each ST encoder comprises two sequentially arranged temporal and spatial Transformers. (\S~\ref{sec:STBlock}).
The skip connections of each ST encoder are summed and fed into an output layer with a transposed 1D convolutional layer(\S~\ref{sec:output}). 
Furthermore, we propose a hierarchical graph structure perception mechanism for spatial attention based on structural entropy optimization to exploit the rich information embedded in road networks.
It abstracts the graph into a hierarchy (i.e., encoding tree), based on which we present multi-level attention masks to regularize spatial attention and hierarchical correlation scores as relative position encoding (\S~\ref{sec:MRSM}).

\subsection{Multi-filter Convolution Module}
\label{sec:MFSTC}

\begin{figure}[t]
    \centering
    \includegraphics[width=0.46\textwidth]{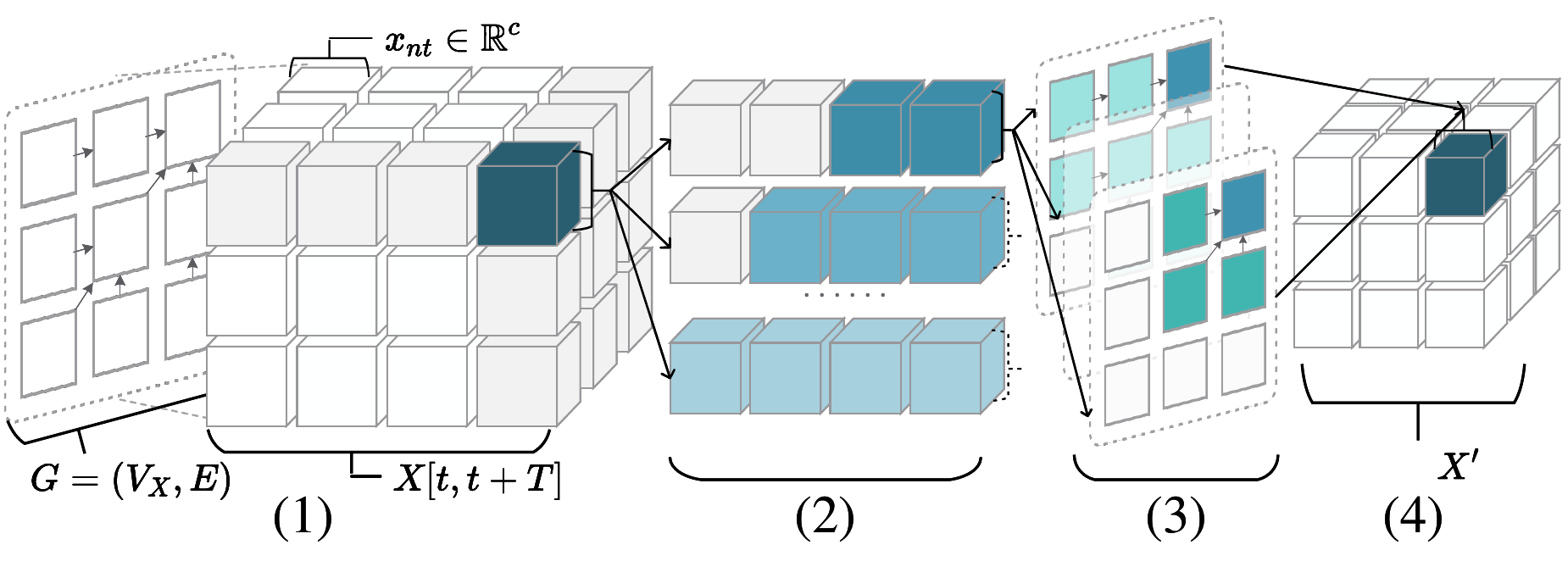}
    \caption{An illustration of the workflow of multi-filter convolution module. (1) 3D spatiotemporal data with the $T$-step time window and a predefined graph $G$. Each atomic data point has $c$-channel attribute; (2) Multiple temporal convolutional filters are employed to extract diverse short-range time patterns; (3) Graph convolutional filters are added for neighborhood aggregation that facilitates the local spatial pattern; (4) Processed data enjoy more extensive channels $c_t$.}
    \label{fig:filter_example}
\end{figure}

%% 分析目的：Transformer从两个维度捕获数据点之间的依赖，单个数据点内数据信息量过少，没有丰富的短期模式信息。ST-Token channel
The multi-filter convolution (MFCL) module aims to expand the dimensionality and enrich the information of token embeddings while incorporating more intricate local spatiotemporal features and patterns. 
We employ two specific designs: multi-frequency temporal convolution filters and multi-hop graph convolution filters.
Fig. ~\ref{fig:filter_example} illustrates the data structure and workflow of this module.

\noindent \textbf{Temporal Convolution Filter}
Recognizing the inherent periodicity of the traffic system, we employ a set of standard 1D filters with various sizes to extract short-range temporal features at multiple frequencies.
Suppose there are $m$ filters with sizes ${k_1,k_2,\cdots,k_m}$, the temporal convolution operation with $c$-channel input and $c_t$-channel output at time $t$ can be formulated as follows: 
% \vspace{-\parskip}
% \begin{align}
%    x'_{t}& = \operatorname{||}_{j=1}^{m}W^{(j)} \ast X[i,t]\\
%    & =\operatorname{||}_{j=1}^{m} \textstyle \sum_{i=1}^{c} \textstyle \sum_{l=1}^{k_j} W^{(j)}[l, i]X[i, t+l-k_j],
% \end{align}
% \vspace{-\parskip}
\begin{equation}
   x'_{t} = \operatorname{||}_{j=1}^{m} \textstyle \sum_{i=1}^{c} \textstyle \sum_{l=1}^{k_j} W^{(j)}[l, i]X[i, t+l-k_j],
\end{equation}
where $X \in \mathbb{R} ^ {T \times c}$ is input time series, $x'_t\in \mathbb{R}^{c_t}$ denotes the output at step $t$, and $W^{(j)} \in \mathbb{R}^{k_j \times c\times  (c_t/m)}$ is the kernel matrix of $j$-th filter (where $c_t$ must be divisible by $m$). 
$\left [ \cdot \right ]$ is the index operation, and $\operatorname{||}$ is the concatenation operation along the channel dimension. 
By concatenating the multiple filters' results, the channel of temporal data is extended to $c_t$.
Since all filters are expected to produce sequences of a uniform length, we padded the sequence to $T \gets T+k_j-1$ in length by duplicating the first and last point of the sequence before feeding into the $j$-th filter.
% Moreover, since each data point aggregates short time series patterns, we can compress the output sequence by increasing the stride $s$ of the convolution through $T_o = \lceil\frac{T}{s}\rceil$.
The size and number of convolution filters can be customized for different tasks to accommodate larger historical windows, and the uniform stride of temporal filters can be enlarged to compress the sequence.
Our basic implementation selects four filters with size $1\times1$, $1\times2$, $1\times3$, and $1\times6$, often corresponding to intervals of 5, 10, 15, and 30 minutes.

\noindent \textbf{Graph Convolution Filter}
% Temporal influence shifts rapidly on the road network, e.g.,  nodes on the same road having similar states or congestion at one intersection soon impact the traffic flow at adjacent intersections. 
To extract the short-range spatial pattern of the traffic state that propagates on the road network, multi-hop graph convolution filters are adopted to fuse the node feature within the neighborhood.
% To fit diverse road network structures, we employ multi-hop neighborhood convolution kernels whose parameters are determined by the edge weights. 
% When the distance between two nodes is greater than a certain threshold, the corresponding entry in the adjacency matrix is marked as 1; otherwise, it is marked as 0.
Denoting the $1$-hop adjacency matrix of the graph as $A$, the $h$-hop graph convolution operation with $c_t$-channel can be formulated as:
\begin{equation}
   x'_{n} = \operatorname{||}_{j=0}^{h} \textstyle \textstyle \sum_{i=1}^{N} \hat{A}^{j}[n, i]X[I],\hat{A} = D^{-1}(A+I).
   % W^{(j)}.
\end{equation}
% \begin{equation}
%  X'_{out}=\operatorname{||}_{j=0}^{h}H^{(j)}
%        =\operatorname{||}_{j=0}^{h}\hat{A}^jX
%    % W^{(j)}.
% \end{equation}
% where
% \begin{equation}
%    \hat{A} = \hat{D}^{-1}(A+I)
% \end{equation}
% \begin{equation}
%    \hat{D}_{ii} = \mathrm{1} + \textstyle \sum_j{A_{ij}}
% \end{equation}
The kernel matrix $\hat{A}$ is derived by adding the self-loop matrix $I$ to $A$ and normalizing it with the degree matrix $D$. 
$X \in \mathbb{R} ^ {N \times c_t}$ denotes the node features with $c_t$-channel at the moment $t$ after temporal filtering, and $x'_{n}$ is the output of the $n$-th node.
$\hat{A}^{j}$ refers to the $j$-th power of $\hat{A}$, which acts as the multi-hop graph convolution filter that aggregation messages from $j$-hop neighbors. 
Finally, all the outputs are concatenated into a vector of dimension $d = (h+1) \cdot c_t$, allowing each data point to aggregate the multi-hop neighborhood locality of the road network.
Our method aggregates the neighborhood representation of each hop independently and has fewer trainable parameters than other similar designs~\cite{abu2019mixhop,wuConnectingDotsMultivariate2020}.

\subsection{Spatial-Temporal Transformer Network}
\label{sec:STBlock}
\subsubsection{Position Embedding Layer}
\label{sec:pe}
% Unlike GNN and RNN networks, the attention mechanism models a wide range of pairwise relations, while its structure prevents it from directly capturing positional information among tokens. Consequently, spatiotemporal tasks necessitate using positional embeddings for sequences and graphs.
% the attention aims to model the wide pair-wise relations while cannot directly capture the position information among tokens. Therefore, sequence and graph position embedding is required in spatiotemporal tasks.

First, to integrate the spatial node position within the spatial Transformer, we utilize the Laplacian graph matrix to encode the road network topology into static representations ~\cite{DwivediTransformertoGraphs2021}. Specifically, We compute the node eigenvectors of $G$ via $U^T\Lambda U = I - D^{-1/2} A D^{-1/2}$, where $U$ and $\Lambda$ correspond to eigenvalues and eigenvectors. A linear projection $W \in \mathbb{R}^{k \times d}$ is applied on $k$ smallest non-trivial eigenvectors to generate the spatial embedding $\mathcal{D}_s \in n \times d$. 
Second, we employ the \textit{Sinusoidal} position encoding $\mathcal{D}_t \in t \times d$ based on the original Transformer ~\cite{VaswaniAttention2017} design to incorporate temporal sequential information.
In addition, for continuous time-series datasets, the position of the current batch within the entire dataset needs to be considered.
We perform one-hot encoding on the day-of-week and hour-of-day timestamps of the data batch and map them into $\mathcal{D}_b \in t \times d$ to account for cross-batch periodicity.
Finally, $H + \mathcal{D}_t + \mathcal{D}_b$ and $H + \mathcal{D}_s + \mathcal{D}_b$ are fed into the spatial and temporal Transformer, respectively. Here, $H \in \mathbb{R}^{ T \times N \times d}$ is the hidden output state of the previous module due to sequential arrangement.

\subsubsection{Spatial-Temporal Transformer}
\label{sec:STTrans}

In order to model global spatiotemporal dependencies on global road networks and historical windows, we employ the unified transformer module with $h$-head attention, which can be formulated as:
\begin{equation}
    % Q = W^{(Q)} (H+\mathcal{D}+\mathcal{D}_b ),K = W^{(K)} H,V = W^{(V)} H,
    Q = W_Q^{(i)}(H+\mathcal{D}+\mathcal{D}_b ),K = W_Q^{(i)} H,V = W_V^{(i)} H,\\
\end{equation}
\begin{equation}
    A^{(i)} = ((Q^{(i)}\cdot K^{(i)})^T+S/\sqrt{d})\odot M^{(i)},\\
\end{equation}
\begin{equation}
    H' = \operatorname{Norm}(\operatorname{RelU}(W_{ffn}\cdot \operatorname{||}^h_{i=1}(\operatorname{softmax}(A^{(i)})V^{(i)}))).
\end{equation}
For the $i$-th attention head, $H$ is the spatiotemporal input (as Fig. ~\ref{fig:filter_example} shows) and $W^{(i)}_Q$,$W^{(i)}_K$, and $W^{(i)}_V$ are learnable linear projection.
$A^{(i)}$ is the attention matrix, $S$ is an addition similarity matrix (also denoted as relative position encoding), and $M^{(i)}$ is the attention mask, which is Hadamard product ($\odot$) with $A^{(i)}$. The final outputs $H'$ of all heads are concatenated into $d$-channel and further fed into a channel-mixing feed-forward layer, where $W_{ffn}$ is the parameter of the feed-forward network and $\operatorname{Norm}$ is the batch normalization. 
The structure of the Temporal and Spatial Transformer is basically the same, but there are still the following differences: 1. the position encodings $D$ are differently obtained. As described in ~\S ~\ref{sec:pe}, 
$D$ in Temporal Transformer is $D_t$, while it is $D_s$ in Spatial Transformer; 2. Temporal Transformer only models the relationship between time points, with all spatial locations share a set of projection parameters, while the opposite is the case for spatial Transformer; 3. We design a unique relative location encoding $S$ and a multi-head attention mask $M$ for spatial attention in ~\S ~\ref{sec:MRSM}.

\begin{figure}[t]
    \centering
    \includegraphics[width=0.48\textwidth]{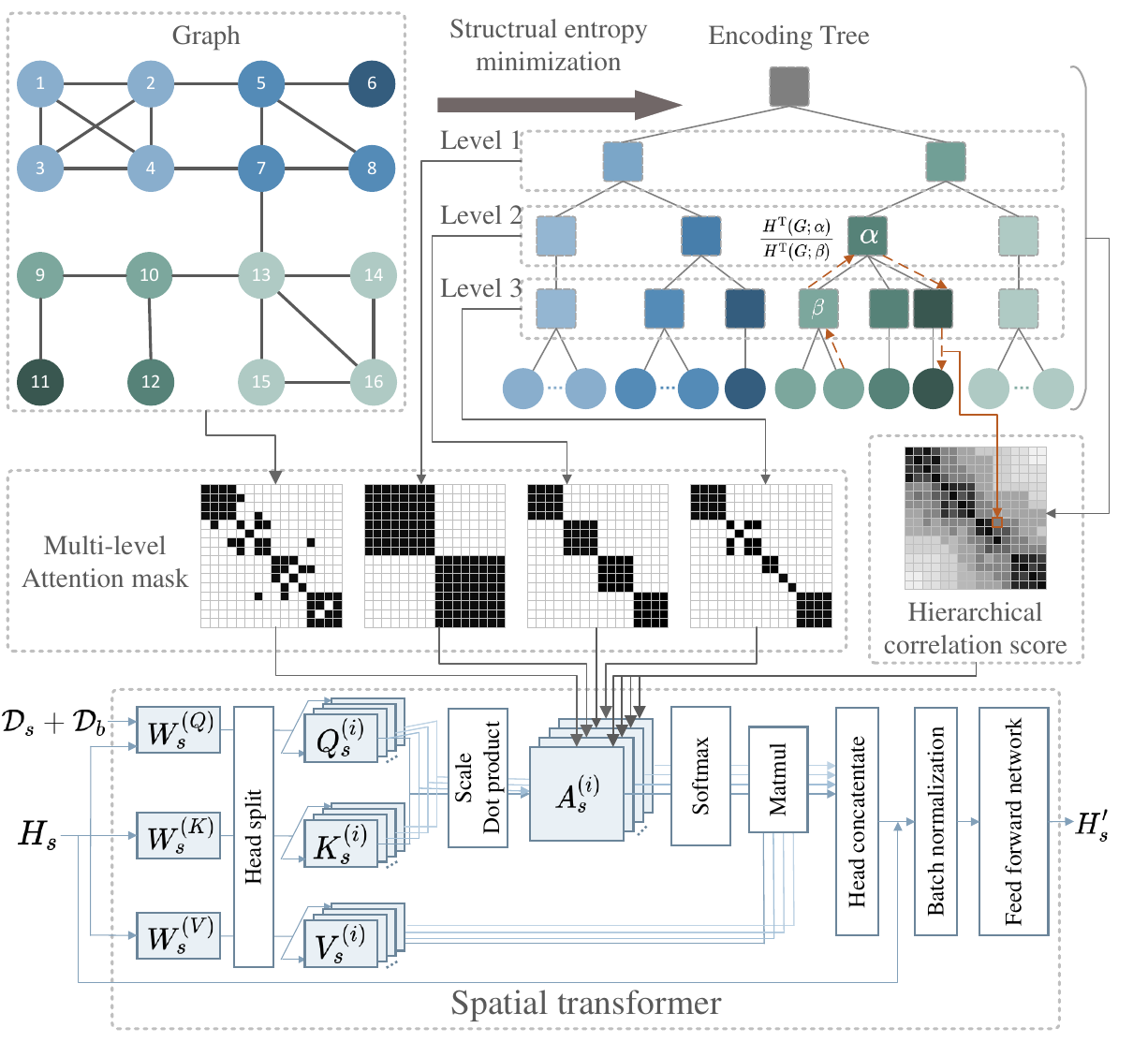}
    \caption{An illustration depicting the hierarchical graph perception mechanism and the spatial Transformer.}
    \label{fig:spatial_transformer}
\end{figure}

\subsubsection{Multi-Range Graph Structure Perception}
\label{sec:MRSM}
The urban fabric has a natural hierarchy due to its functional division (e.g. residential, commercial, etc.), which can be reflected by the road network structure and influence the traffic state.
Structural entropy and encoding tree theory are innovatively introduced to mine higher-order knowledge from the road network and incorporate it into the self-attention mechanism.
Firstly, we apply the structural entropy minimization algorithm to obtain an optimal encoding tree, which serves as a hierarchical abstraction of the road network. 
Secondly, we use the hierarchy to model the low-rank relationship within the network and propose multi-level attention masks. 
Finally, we propose the hierarchical correlation score based on the relative position of physical (leaf) nodes on the encoding tree, which reveals the road network's underlying structure and node positions.

\noindent \textbf{Road Network Abstraction}
Drawing inspiration from the principle of structural entropy minimization~\cite{li2016structural},  we introduce a heuristic algorithm and corresponding tree operators (i.e., the combination operator and merge operator) from deDoc~\cite{li2018decoding} to compute the optimal encoding tree of road network $G$ to obtain a hierarchical zoning structure.
First, we initial a flat encoding tree (with only one level where all leaf nodes are direct descendants of the root node). 
% Then the combination ($\operatorname{CB}$) and merge ($\operatorname{MG}$) operators are greedily conducted on node pair of encoding tree. 
For each iteration, the node pair and operator that maximize the reduction of structural entropy are selected and conducted in a greedy manner.
% Each iteration seeks to maximize the reduction in structural entropy $\delta SE$.
In the end, the algorithm terminates when the structural entropy ceases to decrease continuously, resulting in the final optimal encoding tree denoted as $\mathrm{T}^*$.

\noindent \textbf{Multi-level Attention Mask}
The number of levels in an encoding tree generally depends on the size of the graph and its structural complexity and can be determined adaptively during the optimization.
Each level of the encoding tree corresponds to a partition of the graph node-set, representing the road network potential zoning at a specific spatial scale.
Given $\{\alpha_1,\alpha_2,\dots,\alpha_n\}$ on the $l$-th level of the optimal encoding tree $\mathrm{T}^*$ and $\mathcal{T}_\lambda = {\textstyle \bigcup_{i=1}^{l}} \mathcal{T}_{\mathrm{\alpha}_i}$, we can acquire the mask matrix $M^{(l)} \in \{-INF,1\}^{N \times N}$ that satisfied
\begin{equation}
   m_l[i,j] =
    \begin{cases}
     1 & \text{ if }\exists {\alpha _m}\in\{\alpha_1,\dots,\alpha_l\},  v_i \in \mathcal{T}_{\alpha _m},  v_j \in \mathcal{T}_{\alpha _m}\\ 
     -INF & \text{ else }
    \end{cases},
\end{equation}
where $m_l[i,j]$ denotes the element in the $i$-th row and $j$-th column of $M_l$.  For an $L$-level encoding tree, we can obtain $L-1$ mask matrices with diverse granularity from every level except for the leaf level. 
In addition, we introduce an additional adjacency matrix as the $L$-th mask to capture edge-level local relations with the minimum range.
The $L$ masks are applied to the $H$ attention heads (ensuring $H > L$) to capture dependencies within different ranges, whereas the extra $H-L$ attention heads are unmasked to model the wide global attention.

\noindent \textbf{Hierarchical Correlation Score}
The multi-level attention mask can leverage low-rank constraints on multi-head spatial attention within structural levels but may ignore vertical cross-level relations in hierarchies.
Therefore, we design a relative position encoding to identify vertices in graphs based on the optimal encoding trees.
Specifically, we define the relative structural entropy based on the encoding tree $T$. For nodes $\alpha$ and $\beta$ that have an inheritance relationship, the structural entropy of $\alpha$ relative to $\beta$ is defined as $H^\mathrm{T}_{rel}(G;\alpha|\beta)$=$H^\mathrm{T}(G;\alpha)/H^\mathrm{T}(G;\alpha)$.
It reflects the relative complexity and informativeness between the vertices and sub-structures of the graph $G$.
Then, assuming two leaf nodes $\alpha_i$ and $\alpha_j$ of the encoding tree share the lowest common ancestor $\theta$, the structural entropy of $\alpha_i$ relative to $\alpha_j$ can be defined as follows: 
\begin{equation}
\begin{split}
    H_{rel}^{\mathrm{T}}(G;\alpha_j|\alpha_i)=H_{rel}^{\mathrm{T}}(G;\theta|\alpha_i)+H_{rel}^{\mathrm{T}}(G;\alpha_j|\theta) = \\
    \sum_{\beta,\mathcal{T}_{\alpha_i} \subseteq \mathcal{T}_\beta \subset \mathcal{T}_\theta}
    {H^{\mathrm{T}}(G;\beta^-|\beta )}
    +\sum_{\beta,\mathcal{T}_\theta  \supset \mathcal{T}_\beta \supseteq \mathcal{T}_{\alpha_j}}
    {H^{\mathrm{T}}(G;\beta|\beta^- )}.
\end{split}
\end{equation}
From another perspective, we view the encoding tree as a graph and add up the relative structural entropy of the connected nodes on the shortest directed path between two leaf nodes $\alpha_j,\alpha_i$ to obtain the final relative structural entropy, based on which can we generate the hierarchical correlation matrix satisfying that $S_{hier}[i,j]=H_{rel}^{\mathrm{T}}(G;\alpha_j|\alpha_i)$ where $\mathcal{T}_{\alpha_i} = v_i$ and $\mathcal{T}_{\alpha_j} = v_j$. 
The hierarchical correlation score $S_{hier}$ enables attention to prioritize more intricate structures while preserving the hierarchical information of the road network.
% allows attention to focus on more complex structures while retaining hierarchical information of the road network.
In conclusion, in order to improve the mechanisms of spatial attention, the road network is first abstracted into an encoding tree via the structural entropy minimization algorithm. Then each level $i$ of the encoding tree (and the adjacency matrix) is constructed as an attention mask $M^{(i)}$ that operates on a specific attention head. Furthermore, a hierarchical correlation score $S_{hier}$ derived from relative structural entropy is employed as a prior score and is added to attention matrices.
The modified Spatial Transformer module is depicted in Fig.\ref{fig:spatial_transformer}.

\subsection{Output Layer}
\label{sec:output}
After collecting all intermediate outputs of the ST encoder blocks and the multi-filter convolution filter with the skip connections, they are summed into $H_o \in \mathbb{R}^{T \times N \times D}$ and fed into a deconvolution decoder and an MLP decoder. 
The deconvolution smoothly extends the predicted sequence if the dimension of the hidden states length $T$ is inconsistent with the multi-step predicted length $T'$.
The MLP projects the output's channel dimension and sequence length to the desired shape and obtains the final prediction $H_o \in \mathbb{R}^{T' \times N \times C_o}$.

\section{Results and Analysis}

% This section evaluates \framework{}'s performance on various real-world graph-based traffic datasets with baseline comparison, long-range modelling experiment, ablation study and visual case study. 

% 我们将结果放在表x和表x中
\begin{table*}[th] % 设置跨越双栏的表格
\centering % 表格居中
\caption{Experiment Results of the average 12-step forecast. \textmd{The best results are bolded, and the runner-up results are underlined. \textit{Our} indicates the performance of our purposed \framework{}. \textit{Imp.} denotes the improvement of our method over the SOTA method.}}\label{tab:mainexp}
\scalebox{0.93}{
    \setlength{\tabcolsep}{0.8 mm}{
        \begin{tabular}{c|c|cccccccccccc|c|c@{}} % 设置三列的表格
        \toprule
        \multicolumn{2}{c|}{Methods} &  \multirow{2}{*}{ \space VAR  \space} & \multirow{2}{*}{\space \space SVR \space \space}  & \multirow{2}{*}{\space \space AE \space \space}  & \multirow{2}{*}{ \space LSTM \space}  & \multirow{2}{*}{TGCN}  & \multirow{2}{*}{DCRNN}  &\multirow{2}{*}{STGCN}  & \multirow{2}{*}{MTGNN}  & \multirow{2}{*}{GWNet}  & \multirow{2}{*}{ASTGCN}  & \multirow{2}{*}{STTN}  & \multirow{2}{*}{GMAN} & \multirow{2}{*}{ \space Our  \space} & \multirow{2}{*}{ \space Imp.  \space } \\
        \cline{1-2}
        Dataset & Metrics & & & & & & & & & & & & & \\
        \cline{1-16}
        \multirow{3}{*}{PEMSD4-flow} 
        & MAE   & 24.98 & 27.45 & 24.59 & 23.80 & 22.88 &22.63& 21.60 & \underline{19.29} & 19.53 & 19.56 & 19.49 & 19.35 & \textbf{19.07} & 1.10\% \\
        & MAPE & 18.24 & 19.83 & 16.48 & 15.78 & 14.52 & 13.97 & 14.68 & 13.54 & \underline{13.41} & 13.91 & 13.78 & 13.57 & \textbf{13.29} & 0.90\% \\
        & RMSE & 38.91 & 40.74 &  37.63 & 35.92 & 34.41 & 34.70 & 34.76 & 31.82 &  31.95 & 32.03 & 31.87 & \underline{31.62} & \textbf{30.46} & 3.30\% \\
         \cline{1-16}
        \multirow{3}{*}{PEMSD8-flow} 
        & MAE    & 27.46 & 32.83 & 20.48 & 19.48 & 18.61 & 18.42 & 17.92 & 15.47 & \underline{15.09} & 15.92 & 15.63 &15.34 & \textbf{14.68} &2.72\% \\
        & MAPE & 16.82 & 15.97 & 13.43 & 14.85 & 11.47 & 11.10 & 11.36 & 10.16 & \textbf{9.63} & 10.66 & 10.46 & 10.22& \underline{9.79} & - \\
        & RMSE & 45.01 & 43.95 & 35.19 & 33.27 & 27.95 & 28.14 & 27.34 & 24.93 & \underline{24.84} & 25.37 & 25.26 & 25.13 & \textbf{23.87} & 4.31\% \\
        \cline{1-16}
        \multirow{3}{*}{PEMSD4-speed} 
         & MAE    & 3.29 & 3.15 & 2.35 & 2.58 & 1.94 & 1.70 & 1.80 & 1.67 & \underline{1.66} & 1.80 & 1.72 & 1.74 & \textbf{1.61} & 3.01\% \\
         & MAPE  & 5.90 & 5.77 & 4.79 & 4.17 & 3.77 & 3.60 & 3.57 & 3.48 & \underline{3.45} & 3.94 & 3.68 & 3.64 & \textbf{3.39} & 1.74\% \\
         & RMSE & 5.72 & 6.02 & 4.98 & 5.07 & 4.18 & 3.95 & 3.02 & 3.76 & \underline{3.71} & 3.97 & 3.72 & 3.72 &  \textbf{3.66} & 1.35\% \\
        \cline{1-16}
        \multirow{3}{*}{PEMSD8-speed} 
        & MAE    & 3.14 & 3.60 & 2.13 & 2.35 & 1.73 & 1.51 & 1.55 & 1.47 & \underline{1.42} & 1.59 & 1.54 & 1.49 & \textbf{1.36} & 4.23\% \\
        & MAPE  & 6.39 & 6.48 & 5.04 & 4.96 & 3.42 & 3.26 & 3.28 & \underline{2.95} & 3.06 & 3.62 & 3.61 & 3.41 & \textbf{2.84} & 3.73\% \\
        & RMSE  & 6.83 & 6.13 & 5.35 & 5.29 & 3.67 & 3.64 & 3.50 & 3.49 & 3.56 & 3.73 & 3.90 & \underline{3.43} & \textbf{3.26} & 4.96\% \\
        \bottomrule
        \end{tabular}
        
    }
}
\end{table*}

% We provide evidence of the effectiveness of the proposed multi-filter convolution module and the hierarchical graph perception mechanism.
% A visual case study from spatial and temporal perspectives explains the temporal and spatial dependencies captured by our model.
% The hyperparameter analysis also reveals the robustness of our model while informing the critical hyperparameter selection.

\subsection{Experimental Settings}
% 数据集划分；NVIDIA GeForce 3090; python 3.9 pytorch==1.12.0+cu116;
% , $\beta_1$ = 0.9, and $\beta_2$=0.999
% setting: batch_size 32;
\noindent {\textbf{Implementation.}}
All experiments were performed on the NVIDIA GeForce 3090 with 24GB of memory.
The model was trained by Adam optimizer~\cite{loshchilovFixingWeightDecay2023} with a mean absolute loss (MAE) for $50$ epochs, employing the learning rate $1e-2$ and batch size $32$. The datasets were partitioned into training, validation, and test sets with a ratio of $6:2:2$. The model with the best validation performance was selected for testing.
For a fair comparison, we uniformly configured the number of ST layers as $k=3$, the hidden dimension as $d=64$, and the heads number in self-attention as $h=8$ for all baselines.
% In multi-filter convolution, we set the size of four filters as 1, 2, 3 and 6, respectively, and the maximum number of hops in mix-hop as 4. 

% \subsection{Evaluation Protocols}
\noindent {\textbf{Datasets.}}
% \subsubsection{Datasets}
We conduct experiments on traffic dataset \textbf{PEMSD4}~\cite{guoAttentionBasedSpatialTemporal2019b} and \textbf{PEMSD8}~\cite{guoAttentionBasedSpatialTemporal2019b}.
% We conduct experiments on the following traffic datasets:
% \textbf{(1)PEMSD4}~\cite{guoAttentionBasedSpatialTemporal2019b}
% %: This traffic dataset comprises traffic flow, speed, and occupancy data collected from San Francisco, USA. The traffic road graph consists of 307 nodes and 340 edges. The dataset spans 16,692 time steps, with a time interval of 5 minutes.
% \textbf{(2) PEMSD8}~\cite{guoAttentionBasedSpatialTemporal2019b}
% %: This traffic dataset includes flow, speed, and occupancy information from California highways. The traffic road graph is composed of 170 nodes and 295 edges. The dataset covers 17,856 time steps, with a time interval of 5 minutes.
Both include flow, speed, and occupancy information, with an interval of 5 minutes.
We use all channels as input and select one as the output, based on which we derive four subsets: PEMSD4-speed, PEMSD4-flow, PEMSD8-speed, and PEMSD8-flow.

\noindent {\textbf{Evaluation Metrics.}}
% \subsubsection{Evaluation Metrics}
Three metrics, mean absolute error (MAE), mean absolute percentage error (MAPE), and root mean square error (RMSE), are used for evaluation. 
% When calculating these metrics, missing values in the data are masked. 
Additionally, the average error of output steps was reported to evaluate comprehensively.

\noindent {\textbf{Baselines.}}
% \subsubsection{Baselines}
We compare ~\framework{} against the following baseline methods of four types.
\textbf{Traditional Methods}: Models that apply traditional machine learning methods, including Support Vector Regression (SVR)~\cite{druckerSupportVectorRegression1996a} and Vector Auto Regression(VAR)~\cite{luIntegratingGrangerCausality2016};
\textbf{Deep Learning Methods}: Methods that apply deep approaches excluding GNN or attention, including AutoEncoder(AE)~\cite{lvTrafficFlowPrediction2015} and LSTM~\cite{hochreiterLongShortTermMemory1997};
\textbf{Advanced Methods}: Model specialized for spatiotemporal traffic data with a subtle combination of TCN/RNN and GNN, including TGCN~\cite{zhaoTGCNTemporalGraph2020}, STGCN~\cite{yuSpatioTemporalGraphConvolutional2018}, MTGNN~\cite{wuConnectingDotsMultivariate2020}, and GWNET~\cite{wuGraphWaveNetDeep2019};
\textbf{Transformer-based Methods}: Methods using attention to capture both spatial and temporal dependencies, including ASTGCN~\cite{guoAttentionBasedSpatialTemporal2019}, STTN~\cite{xuSpatialTemporalTransformerNetworks2021}, and GMAN~\cite{zhengGMAN2020};
Implementation of the baselines comes from the Libcity\footnote{https://libcity.ai/\#/}~\cite{WangLibCity2021} benchmark and is adapted to our settings.

\subsection{Experimental Result}

\subsubsection {Comparison with baselines}

A comprehensive comparison between the \framework{} and the baselines is conducted, and the results are reported in Table~\ref{tab:mainexp}.
% It is evident that all deep learning-based approaches outperform traditional ones in traffic forecast, and further improvements can be achieved by explicitly modelling spatial relationships using GNN or Transformer. 
Evidently, all deep learning-based approaches outperform traditional ones in traffic forecast, and further improvements can be achieved by introducing and improving GNN or Transformer for better spatiality. 
% This superiority can be attributed to the strong fitting capability of deep methods for temporal dependencies and the criticality of road network structure. 
We observed that Transformer-based methods generally perform better than GNN-RNN (e.g., STGCN and DCRNN) methods due to their stronger ability to capture global and dynamic dependencies. 
However, MTGNN and GWNET, based on TCN and GNN, show competitive performance and even outperform Transformer-based methods. % on some datasets. 
This may be attributed to their adaptive graph structure learning modules.
% From a temporal perspective, it indicates that leveraging temporal convolution to extract temporal patterns is effective, whereas neglecting this dependency adversely affects the performance of temporal Transformers.
The \framework{} exhibits remarkable performance superiority over baseline methods across all datasets.
Compared to the SOTAs, \framework{} achieves an average improvement of 2.57\%, 2.16\%, and 3.78\% for MAE, MAPE, and RMSE, respectively.
Particularly, ~\framework{} achieves the most significant improvement on PEMSD8-speed, which delivers impressive results of MAE 1.36, MAPE 2.84, and RMSE 3.26, corresponding to the improvements of 4.23\%, 3.73\%, and 4.96\%, respectively.
Additionally, We found that ~\framework{} performs exceptionally well in RMSE, with 23.87 in PESMD8-flow and 30.46 in PESMD4-flow, which may be attributed to the smoothing and denoising impact of the MFCL module and transposed convolutional output layer. 
% These notable promotion demostrate effective utilization of prior knowledge through structure entropy-based hierarchical masks and scores.
% Additionally, the model performs exceptionally well in RMSE, which can be attributed to the effectiveness of the multi-filter convolution module. 
% This module facilitates smoother output and effectively mitigates the impact of outlier data points.

\begin{table}[t]
    \caption{Results with longer windows on PESMD4-flow. }
    
     \begin{tabular}{c|ccc|cc}
    \toprule
     Model & MAE & MAPE & RMSE & Paras. & Time \\
     \midrule
     \framework{}$^2_{-I48}$ & \textbf{18.85} & 13.19 & \textbf{30.18} & 332.3K & 269.15s\\
     \framework{}$^2_{-I36}$ & 18.93 & \textbf{13.17} & 30.25 & 332.3K & 269.48s\\
     \framework{}$^1_{-I48}$ & 19.06 & 13.21 & 30.33 & 332.0K & 266.39s\\
     \framework{}$^1_{-I36}$ & 19.01 & 13.24 & 30.28 & 332.0K &  266.19s\\
     \framework{}$^1_{-I12}$ & 19.07 & 13.29 & 30.46 & 332.0K & 259.46s\\
     \hline
     STTN$_{-I48}$ & 19.31 & 13.55 & 31.74 & 699.8K & 931.18s\\
     STTN$_{-I36}$ & 19.40 & 13.62 & 31.69 & 700.1K & 693.41s\\
     STTN$_{-I12}$ & 19.49 & 13.78 & 31.87 & 700.2K & 178.64s\\
     \hline
     STGCN$_{-I48}$ & 20.97 & 14.42 & 33.35 & 1565.5K & 62.72s\\
     STGCN$_{-I36}$ & 21.31 & 14.45 & 33.44 & 1172.3K & 43.95s\\
     STGCN$_{-I12}$ & 21.60 & 14.68 & 34.76 & 385.9K & 15.56s\\
     \bottomrule
     \end{tabular}
         \label{tab:longseries}
\end{table}

\subsubsection {Long time-series modelling experiments}

\begin{figure*}[t]
    \centering
    \includegraphics[width=0.98\textwidth]{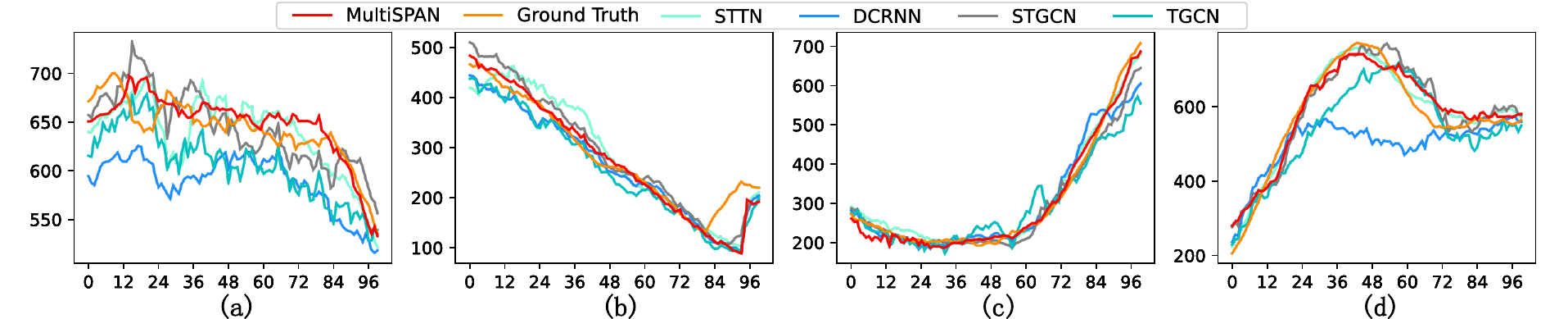}
    \caption{ 
    Forecast results for different periods at the same location. \textmd{We visualized the traffic flow over 100 consecutive time steps using the average results of multiple 12-step forecasts and the ground truth.} 
    % Since we perform multi-step prediction using 12 past steps to predict 12 future steps, there are 12 predicted results for each time point. So we use the average of these predictions as results. To reflect the absolute value of the data and the trend of its change, we show the prediction results for 100 consecutive time steps.
    }
    \label{fig:temporal_variation}
\end{figure*}

In this subsection, we explore the ability of \framework{} to model larger historical time windows and choose a convolution-based (i.e., STGCN) and a transformer-based approach (i.e., STTN) for comparison. 

% As the convolution filter module we designed can compress the temporal dimension of the hidden layer by adjusting the stride, it can model longer-range historical periods.
We adopt the stride of $1,3,4$ for the $12,36,48$ steps historical window for a uniform $12$- length hidden state in \framework{}. In Table \ref{tab:longseries}, $I48,I36,I12$ represent using historical windows of length $48,36,12$. 
\framework{}$^1$ denotes the \framework{} with original settings, while \framework{}$^2$ denotes it with 8 temporal filters of size $[1,2,3,4,6,12,18,24]$.
\textit{Paras.} reports models' total parameter numbers. \textit{Time} reports the average time cost of an epoch. 
The best results are bolded.
As can be observed in Table. ~\ref{tab:longseries}, expanding the history window can improve the performance in most cases, but the extra time and space cost varies among the methods.
% In particular, the improvement in STTN is disproportionate to its incremental time consumption, mainly due to the larger temporal attention matrices associated with its increasing temporal length. 
In particular, the improvement in STTN is disproportionate to its incremental time consumption, mainly due to the increasing computation in dynamic spatial attention on more time patches.
% In particular, the improvement for STTN is insignificant, but the time consumption is considerably increased, proving that the original Transformer is not suitable for real-time long-time series prediction
% In particular, the improvement for TGCLSTM is insignificant, but the time consumption is considerably increased, proving that RNNs are unsuitable for long-range temporal features.
Meanwhile, STGCN improved significantly with longer historical windows, possibly owing to the notable increase in learnable parameters, which also require larger memory.
However, the proposed \framework{} can compress the hidden temporal dimension by tuning the stride of the temporal convolutional filters, thus allowing longer-range history windows to be exploited for improved forecast results at trivial additional cost.
% However, Our proposed \framework{} can model long-term dependencies by self-attention and locality using multi-filter convolution, improving the predictive capability substantially with a limited computational cost. 
Furthermore, extending the number of temporal filters to extract more frequencies of short-range patterns can considerably improve the performance of MultiSPANS to model long-range with a MAE of 18.85, MAPE of 13.17, and RMSE of 30.18.
% Also, extending the convolution filter to extract more frequent short-range patterns can improve modeling long- and short-term dependencies.

\subsection{Ablation Studies}
In this subsection, we conduct an ablation study on the PEMSD4-flow dataset by removing specific modules to evaluate their effectiveness, and results are presented in Table ~\ref{tab:ablationstudy}.

To thoroughly evaluate the multi-filter convolution (MFCL) module, we perform three experiments: 
(1) removing the temporal filter (w\textbackslash o TF), 
(2) removing the spatial filter (w\textbackslash o SF), and 
(3) replacing the MFCL module with a linear layer(w\textbackslash o MFCL). 
It is evident that the improvement of MFCL is dramatic, reaching a surprising 5.24\%. 
Meanwhile, the temporal filter is more effective than the spatial, contributing a 1.98\% improvement compared to 1.49\%.
This observation highlights the necessity of the multi-filter convolution module to extract local patterns for the long-range attention mechanism.

To evaluate the effectiveness of the hierarchical graph perception mechanism, we design experiments to remove or modify its components. 
Specifically, we  
(1) remove the multi-level attention mask(w\textbackslash o mask), 
(2) remove the hierarchical correlation score(w\textbackslash o score), 
(3) remove the whole mechanism(w\textbackslash o mask), and
(4) use the Infomap~\cite{RosvallMapsRandm2008} algorithm, a minimum entropy-based hierarchical community detection method, to construct the multi-level mask(w Infomap). 
% The results show that both the multi-level attention mask and hierarchical correlation score significantly improve the model's performance. The multi-level attention mask leads to a 2.33\% improvement, while the hierarchical correlation score contributes to a 1.52\% improvement, resulting in a total improvement of 4.55\%.
The results show that both the multi-level attention mask and hierarchical correlation score significantly improve the model's performance, contributing to a 2.33\% and a 1.52\% improvement, respectively. And the total improvement of the proposed mechanism amounts to 4.55\%, compared to the vanilla attention.
This suggests that our approach efficiently incorporates topological knowledge into the multi-headed attention, effectively capturing spatial dependencies.
Furthermore, our structural entropy-based method outperforms the Infomap-based method, indicating that structural entropy optimization is more suitable for road network hierarchy abstraction.
Overall, these analyses demonstrate that our design effectively supports multi-range spatio-temporal modeling for traffic.

\begin{table}[t]
    \caption{ Effects of different ~\framework{} components.}
    
     \begin{tabular}{c|ccc|c}
    \toprule
     Model & MAE & MAPE & RMSE & Imp.\\
     \midrule
     \framework{} & 19.07 & 13.29 & 30.46 &  - \\
     \cline{1-5}
     w\textbackslash o TF           & 19.49   & 13.56 & 30.98   & 1.98\%\\
     w\textbackslash o SF           & 19.42   & 13.44 &  30.92  & 1.49\%\\
     w\textbackslash o MFCL         & 20.04   & 14.13 & 31.78   & 5.24\% \\
     \cline{1-5}
     w\textbackslash o mask         & 19.48   & 13.79 & 30.79   & 2.33\%\\
     w\textbackslash o bias         & 19.32   & 13.56 & 30.83   & 1.52\%\\
     w\textbackslash o both         & 19.73    & 14.3 & 31.25    & 4.55\%\\
     w Infomap                      & 19.43   & 13.58 & 30.89   & 1.83\%\\
     \bottomrule
     \end{tabular}
         \label{tab:ablationstudy}
\end{table}

\begin{figure*}[t!]
    \centering
    \includegraphics[width=1.02\textwidth]{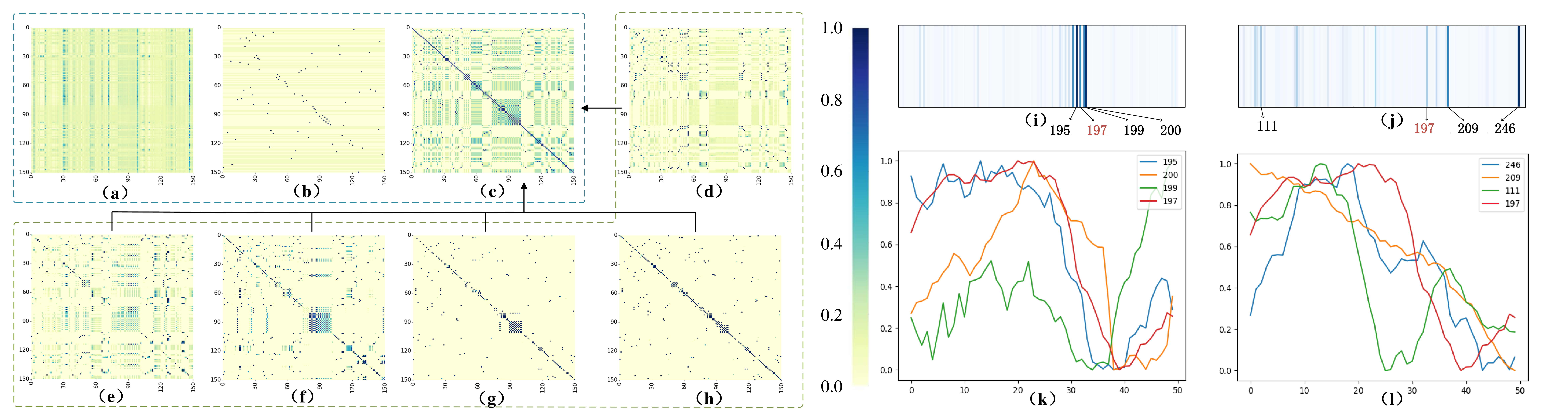}
    \caption{The heatmaps of attention score. 
    \textmd{
    The results of the 100th-250th nodes are shown. 
    (a): The attention map without masks; 
    (b): The attention map masked by adjacency matrix; 
    (c): The average attention map from all heads in \framework{}; 
    (d)$\sim$ (h): The multi-head attention maps with hierarchical graph perception mechanism from coarse to fine granularity; (i)$\sim$(j): The attention heat map between the 197th node and other nodes with vanilla attention and our multi-range method, respectively; (k)$\sim$(l): Current traffic flow at the 197th node and the top 3 relevant nodes based on vanilla attention and our method, respectively.
    }
    }
    \label{fig:attention_score}
\end{figure*}

\subsection{Case Studies}
\subsubsection{Temporal Dependency Study} 
Figure ~\ref{fig:temporal_variation} presents the average prediction of methods in different periods at the same location, along with the corresponding ground truth. 
Specifically, we display the flow prediction of DCRNN, STTN, STGCN, TCN, \framework{}, and the ground truth starting from time steps 72 (a), 432 (b), 792 (c), and 1152 (d) of node 101 in the PEMSD4. 
In (b), (c), and (d), our model's results are smoother and less sensitive to anomalies in comparison. This can be attributed to the denoising effect of the incorporated multi-filter convolution module and temporal deconvolution decoder.
And overall, our model fits the ground truth better, matching trends (b,c) and effectively modeling specific temporal patterns (a,d), indicating its efficiency in temporal modeling.
%(including morning, afternoon, and evening)

% \begin{figure}[t]
%     \centering
%     \includegraphics[width=0.46\textwidth]{figures/line.pdf}
%     \caption{ 
%     Forecast results for different time periods at the same location. 
%     We visualised the results for 100 consecutive time steps using the average results of multiple 12-step predictions.
%     % Since we perform multi-step prediction using 12 past steps to predict 12 future steps, there are 12 predicted results for each time point. So we use the average of these predictions as results. To reflect the absolute value of the data and the trend of its change, we show the prediction results for 100 consecutive time steps.
%     }
%     \label{fig:temporal_variation}

% \end{figure}

\subsubsection{Spatial Dependency Study}
% mask
% score
% 同时，我们也在图中展示模型对空间依赖关系的捕捉结果。首先在图(a),图(b)和图(c)中分别展示了基于结构熵的层次化mask。对应的在图(d)图(e)和图(f)。通过对于结构熵

We also illustrate the spatial attention map captured by ~\framework{} in Fig. ~\ref{fig:attention_score}. 
As shown in Fig. (a), attention is modeled globally without masks, and most nodes rely heavily on a few key nodes in the road network.
Fig. (b) shows the discrete attention matrix when using the adjacency matrix as a mask.
Both attention-modeling approaches drastically lose sight of the complex semantics of the road network.
Meanwhile, as shown in Fig. (a)$\sim$(h), the multi-level attention we designed can capture different range dependencies at each attention head separately.
The fusion of the attention map provided by the hierarchical graph perception mechanism (Fig. (c)) shows that our approach is able to model richer spatiality than vanilla attention. %and graph attention. 
To interpret the plausibility of the attention of our method, we further analyze temporal patterns among closely related nodes.
Specifically, we selected the three nodes with the strongest relevance to 197 points based on multi-level attention (Fig. (i)) and vanilla attention (Fig. (j)) and visualized their corresponding local flow in Fig.(k) and Fig.(l), respectively.
% We observe that the temporal trends of the nodes captured by our method are much closer (e.g., node 195 is almost identical to node 197), which means that our method is more effective in local modeling.
While if no multi-level constraints are added, long-range relationships can be captured (e.g., nodes 246 and 197), but the overall similarity is not pronounced.

% We found that the three most relevant points to point 197 are point 195, point 199 and point 200 when all masks are used, while these are point 111, point 209 and point 246 when no mask is used (Figure(i)$\sim$(j)). 
% It can be seen that point 197 has the same trend as points 195 and  200, while points 199 and 200 have the exact opposite trend. 
% In contrast, the flow trends of points 111, 209, and 246 exhibit limited relevance to point 197. 
% Therefore, we can conclude that the proposed structure entropy-based hierarchical mask can effectively assist the model in extracting spatial dependencies.

\subsubsection{Hyperparameter Analysis}

\begin{figure}[t!]
    \centering
    \includegraphics[width=0.48\textwidth]{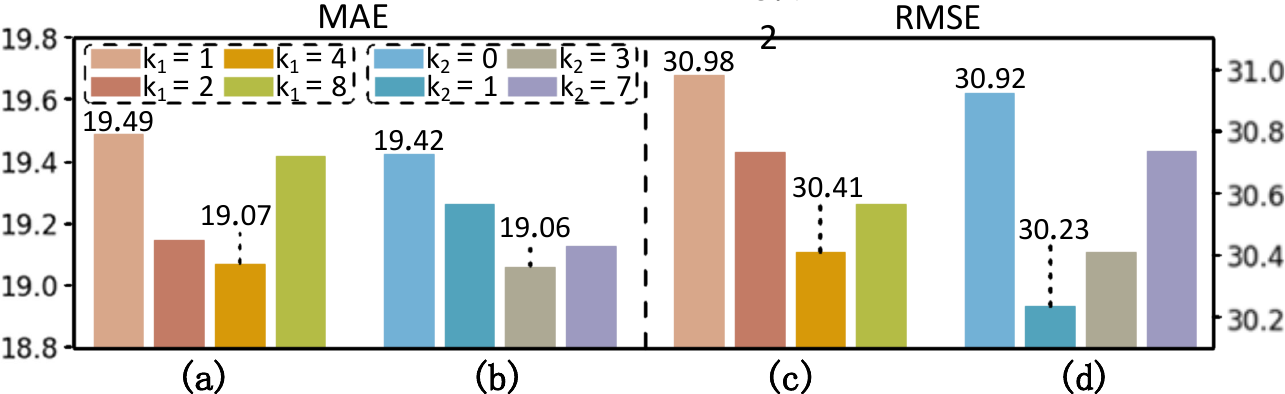}
    \caption{~Influence of hyperparameters. \textmd{Figure (a), (b) shows the influence of $k_1$, and Figure(b), (d) shows that of $k_2$.}}
    \label{hyper1}
\end{figure}

Fig.~\ref{hyper1} evaluates two hyperparameters on PEMSD4-flow, i.e., the temporal filter number $k_1$ and hops of the spatial filter $k_2$.
% The result increases with the number of filters and peaks at $k_1=4$ (Fig. (a)-(c)).
Appropriate $k_1$ and $k_2$ do promote model performance in terms of extracting extensive local patterns and avoiding excessive noise. Meanwhile, they generally remain at a high level and exhibit relative stability, indicating that our method is not sensitive to the hyperparameters.
Further, Noting that even with $k_1=1$ or $k_2=0$ (i.e. with only one temporal filter or not exhibiting spatial neighborhood), \framework{} still achieves RMSEs of 30.98 and 30.92, respectively, exceeding most existing models. 

\section{RELATED WORK}\label{sec:related_work}

\noindent {\textbf{Deep traffic forecast}}
Deep traffic forecast is a spatiotemporal regression task involving GNN, RNN, TCN, and Transformer~\cite{wang2020deep} etc.
Learning spatiality with GNN and predicting with RNN is a typical paradigm ~\cite{zhaoTGCNTemporalGraph2020, CuiTraffic2019, ChenMultiRangeAttentive2020, liDiffusionConvolutionalRecurrent2022,xia2022multi}. Meanwhile, deep convolutional approaches of stacking GNN and TCN modules have also proved effective, which ameliorates the localization problem~\cite{yuSpatioTemporalGraphConvolutional2018,guoHierarchicalGraphConvolution2021,wuGraphWaveNetDeep2019, LuSpatiotemporalAdaptiveGated2020,wuGraphWaveNetDeep2019,guoHierarchicalGraphConvolution2021}.
To further improve the capabilities, some work aims to utilize traffic-related attributes, like hour-of-day and day-of-week etc~\cite{shaoSpatialTemporalIdentitySimple2022,guoAttentionBasedSpatialTemporal2019, yaoRevisitingSpatialTemporalSimilarity2019}, and some adopt graph structure learning for high-quality and task-relevant road network structures~\cite{wuConnectingDotsMultivariate2020, wuGraphWaveNetDeep2019, ShangDiscreteGraphStructure2022, ZhangSpatioTemporalGraphStructure2020,wang2022knowledge}.
Recently, many studies ~\cite{zhou2021informer, wu2021autoformer, liMTSMixersMultivariateTime2023, nieTimeSeriesWorth2023} have endorsed Transformers in long time series, despite some deficiencies ~\cite{zengAreTransformersEffective2022} such as poor information in single tokens ~\cite{nieTimeSeriesWorth2023}.
Therefore, advanced work~\cite{zhengGMAN2020, guoLearningDynamicsHeterogeneity2022, guoAttentionBasedSpatialTemporal2019, ParkSTGRAT2020, xuSpatialTemporalTransformerNetworks2021, jiangPDFormer2023} is keen to model both temporal and spatial dependencies with Transformers.
For example, ASTGNN~\cite{guoLearningDynamicsHeterogeneity2022} propose a dynamic tri-multi-head self-attention, and STTN~\cite{xuSpatialTemporalTransformerNetworks2021} incorporates GCNs and spatial Transformers with the gated-fusion.
PDformer~\cite{jiangPDFormer2023} adopts geographic and semantic spatial masks on attention heads.

\noindent {\textbf{Structural entropy application.}}
To evaluate the quality and informativeness of the graph structure, many works~\cite{RosvallMapsRandm2008,li2016structural, OmarInformationEntropy2020} are presented to extend the Shannon entropy~\cite{shannon1948mathematical} to structural data.
Among which, structural information theory ~\cite{li2016structural}, as a de-facto solution to measure information in graphs, was first applied in network security~\cite{li2017resistance,liu2019rem,li2016resistance} and bioinformatics~\cite{li2016three,li2018decoding,zeng2023unsupervised}, etc.
Recently, a wave of work has been aimed at applying structural entropy to cutting-edge machine-learning areas.  
Some work has attempted to improve GNNs by structural entropy, i.e., selecting optimal hyperparameters~\cite{yang2023minimum}, learning graph structures~\cite{zou2023se}, or designing pooling frameworks~\cite{wu2022structural}.
Some work combines structural information with reinforcement learning to optimize role ~\cite{zeng2023effective} and state~\cite{zeng2023hierarchical} abstraction, with promising results achieved.

\section{CONCLUSION}
We address multi-range spatial modeling from the structural entropy perspective and propose a novel Transformer-based traffic forecast framework.
Consisting of a multi-filter convolution module, road network abstraction, and graph perception mechanism, ~\framework{} succeeds in spatiotemporal tokenizing, discovering road network hierarchy, and poses the multi-level constraint on Transformers.
Experiments show that ~\framework{} achieves excellent performance, and demonstrate the effectiveness of proposed modules.
In the future, we plan to focus on applying structural entropy-guided attention mechanisms to graph and spatial data and analyze the Transformer's interpretability from the hierarchical network analysis perspective.

\begin{acks}
This work is supported by National Key R\&D Program of China through grant 2021YFB1714800, NSFC through grants 62322202 and 62206148, Beijing Natural Science Foundation through grant 4222030, S\&T Program of Hebei through grant 20310101D, China Postdoctoral Science Foundation through grant 2022M711811, CCF-DiDi GAIA Collaborative Research Funds for Young Scholars, and the Fundamental Research Funds for the Central Universities.
\end{acks}

\bibliographystyle{ACM-Reference-Format}
\balance
\bibliography{sample}


\begin{thebibliography}{64}

%%% the Web address.

\ifx \showCODEN    \undefined \def \showCODEN     #1{\unskip}     \fi
\ifx \showDOI      \undefined \def \showDOI       #1{#1}\fi
\ifx \showISBNx    \undefined \def \showISBNx     #1{\unskip}     \fi
\ifx \showISBNxiii \undefined \def \showISBNxiii  #1{\unskip}     \fi
\ifx \showISSN     \undefined \def \showISSN      #1{\unskip}     \fi
\ifx \showLCCN     \undefined \def \showLCCN      #1{\unskip}     \fi
\ifx \shownote     \undefined \def \shownote      #1{#1}          \fi
\ifx \showarticletitle \undefined \def \showarticletitle #1{#1}   \fi
\ifx \showURL      \undefined \def \showURL       {\relax}        \fi
% invisible to TeX
\providecommand\bibfield[2]{#2}
\providecommand\bibinfo[2]{#2}
\providecommand\natexlab[1]{#1}
\providecommand\showeprint[2][]{arXiv:#2}

\bibitem[Abu-El-Haija et~al\mbox{.}(2019)]%
        {abu2019mixhop}
\bibfield{author}{\bibinfo{person}{Sami Abu-El-Haija}, \bibinfo{person}{Bryan Perozzi}, \bibinfo{person}{Amol Kapoor}, \bibinfo{person}{Nazanin Alipourfard}, \bibinfo{person}{Kristina Lerman}, \bibinfo{person}{Hrayr Harutyunyan}, \bibinfo{person}{Greg Ver~Steeg}, {and} \bibinfo{person}{Aram Galstyan}.} \bibinfo{year}{2019}\natexlab{}.
\newblock \showarticletitle{Mixhop: Higher-order graph convolutional architectures via sparsified neighborhood mixing}. In \bibinfo{booktitle}{\emph{international conference on machine learning}}. PMLR, \bibinfo{pages}{21--29}.
\newblock


\bibitem[Bai et~al\mbox{.}(2020)]%
        {BaiAdaptive2020}
\bibfield{author}{\bibinfo{person}{Lei Bai}, \bibinfo{person}{Lina Yao}, \bibinfo{person}{Can Li}, \bibinfo{person}{Xianzhi Wang}, {and} \bibinfo{person}{Can Wang}.} \bibinfo{year}{2020}\natexlab{}.
\newblock \showarticletitle{Adaptive Graph Convolutional Recurrent Network for Traffic Forecasting}.
\newblock \bibinfo{journal}{\emph{Advances in Neural Information Processing Systems}}  \bibinfo{volume}{33} (\bibinfo{year}{2020}), \bibinfo{pages}{17804--17815}.
\newblock


\bibitem[Chen et~al\mbox{.}(2020b)]%
        {chen2020simple}
\bibfield{author}{\bibinfo{person}{Ming Chen}, \bibinfo{person}{Zhewei Wei}, \bibinfo{person}{Zengfeng Huang}, \bibinfo{person}{Bolin Ding}, {and} \bibinfo{person}{Yaliang Li}.} \bibinfo{year}{2020}\natexlab{b}.
\newblock \showarticletitle{Simple and deep graph convolutional networks}. In \bibinfo{booktitle}{\emph{International conference on machine learning}}. PMLR, \bibinfo{pages}{1725--1735}.
\newblock


\bibitem[Chen et~al\mbox{.}(2020a)]%
        {ChenMultiRangeAttentive2020}
\bibfield{author}{\bibinfo{person}{Weiqi Chen}, \bibinfo{person}{Ling Chen}, \bibinfo{person}{Yu Xie}, \bibinfo{person}{Wei Cao}, \bibinfo{person}{Yusong Gao}, {and} \bibinfo{person}{Xiaojie Feng}.} \bibinfo{year}{2020}\natexlab{a}.
\newblock \showarticletitle{Multi-Range Attentive Bicomponent Graph Convolutional Network for Traffic Forecasting}.
\newblock \bibinfo{journal}{\emph{Proceedings of the AAAI Conference on Artificial Intelligence}} \bibinfo{volume}{34}, \bibinfo{number}{04} (\bibinfo{date}{April} \bibinfo{year}{2020}), \bibinfo{pages}{3529--3536}.
\newblock
\showISSN{2374-3468}
\urldef\tempurl%
\url{https://doi.org/10.1609/aaai.v34i04.5758}
\showDOI{\tempurl}


\bibitem[Cui et~al\mbox{.}(2019)]%
        {CuiTraffic2019}
\bibfield{author}{\bibinfo{person}{Zhiyong Cui}, \bibinfo{person}{Kristian Henrickson}, \bibinfo{person}{Ruimin Ke}, {and} \bibinfo{person}{Yinhai Wang}.} \bibinfo{year}{2019}\natexlab{}.
\newblock \showarticletitle{Traffic graph convolutional recurrent neural network: A deep learning framework for network-scale traffic learning and forecasting}.
\newblock \bibinfo{journal}{\emph{IEEE Transactions on Intelligent Transportation Systems}} \bibinfo{volume}{21}, \bibinfo{number}{11} (\bibinfo{year}{2019}), \bibinfo{pages}{4883--4894}.
\newblock


\bibitem[Dosovitskiy et~al\mbox{.}(2021)]%
        {DosovitskiyViT2021}
\bibfield{author}{\bibinfo{person}{Alexey Dosovitskiy}, \bibinfo{person}{Lucas Beyer}, \bibinfo{person}{Alexander Kolesnikov}, \bibinfo{person}{Dirk Weissenborn}, \bibinfo{person}{Xiaohua Zhai}, \bibinfo{person}{Thomas Unterthiner}, \bibinfo{person}{Mostafa Dehghani}, \bibinfo{person}{Matthias Minderer}, \bibinfo{person}{Georg Heigold}, \bibinfo{person}{Sylvain Gelly}, \bibinfo{person}{Jakob Uszkoreit}, {and} \bibinfo{person}{Neil Houlsby}.} \bibinfo{year}{2021}\natexlab{}.
\newblock \showarticletitle{An Image Is Worth 16x16 Words: Transformers for Image Recognition at Scale}. In \bibinfo{booktitle}{\emph{International Conference on Learning Representations}}.
\newblock


\bibitem[Drucker et~al\mbox{.}(1996)]%
        {druckerSupportVectorRegression1996a}
\bibfield{author}{\bibinfo{person}{Harris Drucker}, \bibinfo{person}{Christopher J.~C. Burges}, \bibinfo{person}{Linda Kaufman}, \bibinfo{person}{Alex Smola}, {and} \bibinfo{person}{Vladimir Vapnik}.} \bibinfo{year}{1996}\natexlab{}.
\newblock \showarticletitle{Support Vector Regression Machines}. In \bibinfo{booktitle}{\emph{Advances in Neural Information Processing Systems}}, Vol.~\bibinfo{volume}{9}. \bibinfo{publisher}{MIT Press}.
\newblock


\bibitem[Dwivedi and Bresson(2021)]%
        {DwivediTransformertoGraphs2021}
\bibfield{author}{\bibinfo{person}{Vijay~Prakash Dwivedi} {and} \bibinfo{person}{Xavier Bresson}.} \bibinfo{year}{2021}\natexlab{}.
\newblock \bibinfo{title}{A Generalization of Transformer Networks to Graphs}.
\newblock
\newblock
\urldef\tempurl%
\url{https://doi.org/10.48550/arXiv.2012.09699}
\showDOI{\tempurl}
\showeprint[arxiv]{2012.09699}~[cs]


\bibitem[Guo et~al\mbox{.}(2021)]%
        {guoHierarchicalGraphConvolution2021}
\bibfield{author}{\bibinfo{person}{Kan Guo}, \bibinfo{person}{Yongli Hu}, \bibinfo{person}{Yanfeng Sun}, \bibinfo{person}{Sean Qian}, \bibinfo{person}{Junbin Gao}, {and} \bibinfo{person}{Baocai Yin}.} \bibinfo{year}{2021}\natexlab{}.
\newblock \showarticletitle{Hierarchical Graph Convolution Network for Traffic Forecasting}.
\newblock \bibinfo{journal}{\emph{Proceedings of the AAAI Conference on Artificial Intelligence}} \bibinfo{volume}{35}, \bibinfo{number}{1} (\bibinfo{date}{May} \bibinfo{year}{2021}), \bibinfo{pages}{151--159}.
\newblock
\showISSN{2374-3468}
\urldef\tempurl%
\url{https://doi.org/10.1609/aaai.v35i1.16088}
\showDOI{\tempurl}


\bibitem[Guo et~al\mbox{.}(2019a)]%
        {guoAttentionBasedSpatialTemporal2019}
\bibfield{author}{\bibinfo{person}{Shengnan Guo}, \bibinfo{person}{Youfang Lin}, \bibinfo{person}{Ning Feng}, \bibinfo{person}{Chao Song}, {and} \bibinfo{person}{Huaiyu Wan}.} \bibinfo{year}{2019}\natexlab{a}.
\newblock \showarticletitle{Attention {{Based Spatial-Temporal Graph Convolutional Networks}} for {{Traffic Flow Forecasting}}}.
\newblock \bibinfo{journal}{\emph{Proceedings of the AAAI Conference on Artificial Intelligence}} \bibinfo{volume}{33}, \bibinfo{number}{01} (\bibinfo{date}{July} \bibinfo{year}{2019}), \bibinfo{pages}{922--929}.
\newblock
\showISSN{2374-3468}
\urldef\tempurl%
\url{https://doi.org/10.1609/aaai.v33i01.3301922}
\showDOI{\tempurl}


\bibitem[Guo et~al\mbox{.}(2019b)]%
        {guoAttentionBasedSpatialTemporal2019b}
\bibfield{author}{\bibinfo{person}{Shengnan Guo}, \bibinfo{person}{Youfang Lin}, \bibinfo{person}{Ning Feng}, \bibinfo{person}{Chao Song}, {and} \bibinfo{person}{Huaiyu Wan}.} \bibinfo{year}{2019}\natexlab{b}.
\newblock \showarticletitle{Attention Based Spatial-Temporal Graph Convolutional Networks for Traffic Flow Forecasting}.
\newblock \bibinfo{journal}{\emph{Proceedings of the AAAI Conference on Artificial Intelligence}} \bibinfo{volume}{33}, \bibinfo{number}{01} (\bibinfo{date}{July} \bibinfo{year}{2019}), \bibinfo{pages}{922--929}.
\newblock
\showISSN{2374-3468}
\urldef\tempurl%
\url{https://doi.org/10.1609/aaai.v33i01.3301922}
\showDOI{\tempurl}


\bibitem[Guo et~al\mbox{.}(2022)]%
        {guoLearningDynamicsHeterogeneity2022}
\bibfield{author}{\bibinfo{person}{Shengnan Guo}, \bibinfo{person}{Youfang Lin}, \bibinfo{person}{Huaiyu Wan}, \bibinfo{person}{Xiucheng Li}, {and} \bibinfo{person}{Gao Cong}.} \bibinfo{year}{2022}\natexlab{}.
\newblock \showarticletitle{Learning {{Dynamics}} and {{Heterogeneity}} of {{Spatial-Temporal Graph Data}} for {{Traffic Forecasting}}}.
\newblock \bibinfo{journal}{\emph{IEEE Transactions on Knowledge and Data Engineering}} \bibinfo{volume}{34}, \bibinfo{number}{11} (\bibinfo{date}{Nov.} \bibinfo{year}{2022}), \bibinfo{pages}{5415--5428}.
\newblock
\showISSN{1558-2191}
\urldef\tempurl%
\url{https://doi.org/10.1109/TKDE.2021.3056502}
\showDOI{\tempurl}


\bibitem[Hochreiter and Schmidhuber(1997)]%
        {hochreiterLongShortTermMemory1997}
\bibfield{author}{\bibinfo{person}{Sepp Hochreiter} {and} \bibinfo{person}{J{\"u}rgen Schmidhuber}.} \bibinfo{year}{1997}\natexlab{}.
\newblock \showarticletitle{Long Short-Term Memory}.
\newblock \bibinfo{journal}{\emph{Neural Computation}} \bibinfo{volume}{9}, \bibinfo{number}{8} (\bibinfo{date}{Nov.} \bibinfo{year}{1997}), \bibinfo{pages}{1735--1780}.
\newblock
\showISSN{0899-7667}
\urldef\tempurl%
\url{https://doi.org/10.1162/neco.1997.9.8.1735}
\showDOI{\tempurl}


\bibitem[Huang et~al\mbox{.}(2019)]%
        {huang2019deep}
\bibfield{author}{\bibinfo{person}{Chao Huang}, \bibinfo{person}{Chuxu Zhang}, \bibinfo{person}{Peng Dai}, {and} \bibinfo{person}{Liefeng Bo}.} \bibinfo{year}{2019}\natexlab{}.
\newblock \showarticletitle{Deep dynamic fusion network for traffic accident forecasting}. In \bibinfo{booktitle}{\emph{Proceedings of the 28th ACM international conference on information and knowledge management}}. \bibinfo{pages}{2673--2681}.
\newblock


\bibitem[Jiang et~al\mbox{.}(2023)]%
        {jiangPDFormer2023}
\bibfield{author}{\bibinfo{person}{Jiawei Jiang}, \bibinfo{person}{Chengkai Han}, \bibinfo{person}{Wayne~Xin Zhao}, {and} \bibinfo{person}{Jingyuan Wang}.} \bibinfo{year}{2023}\natexlab{}.
\newblock \bibinfo{booktitle}{\emph{PDFormer: Propagation Delay-Aware Dynamic Long-Range Transformer for Traffic Flow Prediction}}.
\newblock


\bibitem[Li et~al\mbox{.}(2016a)]%
        {li2016resistance}
\bibfield{author}{\bibinfo{person}{Angsheng Li}, \bibinfo{person}{Qifu Hu}, \bibinfo{person}{Jun Liu}, {and} \bibinfo{person}{Yicheng Pan}.} \bibinfo{year}{2016}\natexlab{a}.
\newblock \showarticletitle{Resistance and security index of networks: structural information perspective of network security}.
\newblock \bibinfo{journal}{\emph{Scientific reports}} \bibinfo{volume}{6}, \bibinfo{number}{1} (\bibinfo{year}{2016}), \bibinfo{pages}{26810}.
\newblock


\bibitem[Li and Pan(2016)]%
        {li2016structural}
\bibfield{author}{\bibinfo{person}{Angsheng Li} {and} \bibinfo{person}{Yicheng Pan}.} \bibinfo{year}{2016}\natexlab{}.
\newblock \showarticletitle{Structural information and dynamical complexity of networks}.
\newblock \bibinfo{journal}{\emph{IEEE Transactions on Information Theory}} \bibinfo{volume}{62}, \bibinfo{number}{6} (\bibinfo{year}{2016}), \bibinfo{pages}{3290--3339}.
\newblock


\bibitem[Li et~al\mbox{.}(2016b)]%
        {li2016three}
\bibfield{author}{\bibinfo{person}{Angsheng Li}, \bibinfo{person}{Xianchen Yin}, {and} \bibinfo{person}{Yicheng Pan}.} \bibinfo{year}{2016}\natexlab{b}.
\newblock \showarticletitle{Three-dimensional gene map of cancer cell types: Structural entropy minimisation principle for defining tumour subtypes}.
\newblock \bibinfo{journal}{\emph{Scientific reports}} \bibinfo{volume}{6}, \bibinfo{number}{1} (\bibinfo{year}{2016}), \bibinfo{pages}{1--26}.
\newblock


\bibitem[Li et~al\mbox{.}(2018)]%
        {li2018decoding}
\bibfield{author}{\bibinfo{person}{Angsheng Li}, \bibinfo{person}{Xianchen Yin}, \bibinfo{person}{Bingxiang Xu}, \bibinfo{person}{Danyang Wang}, \bibinfo{person}{Jimin Han}, \bibinfo{person}{Yi Wei}, \bibinfo{person}{Yun Deng}, \bibinfo{person}{Ying Xiong}, {and} \bibinfo{person}{Zhihua Zhang}.} \bibinfo{year}{2018}\natexlab{}.
\newblock \showarticletitle{Decoding topologically associating domains with ultra-low resolution Hi-C data by graph structural entropy}.
\newblock \bibinfo{journal}{\emph{Nature communications}} \bibinfo{volume}{9}, \bibinfo{number}{1} (\bibinfo{year}{2018}), \bibinfo{pages}{1--12}.
\newblock


\bibitem[Li et~al\mbox{.}(2017)]%
        {li2017resistance}
\bibfield{author}{\bibinfo{person}{Angsheng Li}, \bibinfo{person}{Xiaohui Zhang}, {and} \bibinfo{person}{Yicheng Pan}.} \bibinfo{year}{2017}\natexlab{}.
\newblock \showarticletitle{Resistance maximization principle for defending networks against virus attack}.
\newblock \bibinfo{journal}{\emph{Physica A: Statistical Mechanics and its Applications}}  \bibinfo{volume}{466} (\bibinfo{year}{2017}), \bibinfo{pages}{211--223}.
\newblock


\bibitem[Li et~al\mbox{.}(2022)]%
        {liDiffusionConvolutionalRecurrent2022}
\bibfield{author}{\bibinfo{person}{Yaguang Li}, \bibinfo{person}{Rose Yu}, \bibinfo{person}{Cyrus Shahabi}, {and} \bibinfo{person}{Yan Liu}.} \bibinfo{year}{2022}\natexlab{}.
\newblock \showarticletitle{Diffusion {{Convolutional Recurrent Neural Network}}: {{Data-Driven Traffic Forecasting}}}. In \bibinfo{booktitle}{\emph{International {{Conference}} on {{Learning Representations}}}}.
\newblock


\bibitem[Li et~al\mbox{.}(2023)]%
        {liMTSMixersMultivariateTime2023}
\bibfield{author}{\bibinfo{person}{Zhe Li}, \bibinfo{person}{Zhongwen Rao}, \bibinfo{person}{Lujia Pan}, {and} \bibinfo{person}{Zenglin Xu}.} \bibinfo{year}{2023}\natexlab{}.
\newblock \bibinfo{booktitle}{\emph{MTS-Mixers: Multivariate Time Series Forecasting via Factorized Temporal and Channel Mixing}}.
\newblock


\bibitem[Lian et~al\mbox{.}(2020)]%
        {lian2020geography}
\bibfield{author}{\bibinfo{person}{Defu Lian}, \bibinfo{person}{Yongji Wu}, \bibinfo{person}{Yong Ge}, \bibinfo{person}{Xing Xie}, {and} \bibinfo{person}{Enhong Chen}.} \bibinfo{year}{2020}\natexlab{}.
\newblock \showarticletitle{Geography-aware sequential location recommendation}. In \bibinfo{booktitle}{\emph{Proceedings of the 26th ACM SIGKDD international conference on knowledge discovery \& data mining}}. \bibinfo{pages}{2009--2019}.
\newblock


\bibitem[Liu et~al\mbox{.}(2019)]%
        {liu2019rem}
\bibfield{author}{\bibinfo{person}{Yiwei Liu}, \bibinfo{person}{Jiamou Liu}, \bibinfo{person}{Zijian Zhang}, \bibinfo{person}{Liehuang Zhu}, {and} \bibinfo{person}{Angsheng Li}.} \bibinfo{year}{2019}\natexlab{}.
\newblock \showarticletitle{REM: From structural entropy to community structure deception}.
\newblock \bibinfo{journal}{\emph{Advances in Neural Information Processing Systems}}  \bibinfo{volume}{32} (\bibinfo{year}{2019}).
\newblock


\bibitem[Loshchilov and Hutter(2023)]%
        {loshchilovFixingWeightDecay2023}
\bibfield{author}{\bibinfo{person}{Ilya Loshchilov} {and} \bibinfo{person}{Frank Hutter}.} \bibinfo{year}{2023}\natexlab{}.
\newblock \showarticletitle{Fixing Weight Decay Regularization in Adam}.
\newblock  (\bibinfo{date}{May} \bibinfo{year}{2023}).
\newblock


\bibitem[Lu et~al\mbox{.}(2020)]%
        {LuSpatiotemporalAdaptiveGated2020}
\bibfield{author}{\bibinfo{person}{Bin Lu}, \bibinfo{person}{Xiaoying Gan}, \bibinfo{person}{Haiming Jin}, \bibinfo{person}{Luoyi Fu}, {and} \bibinfo{person}{Haisong Zhang}.} \bibinfo{year}{2020}\natexlab{}.
\newblock \showarticletitle{Spatiotemporal Adaptive Gated Graph Convolution Network for Urban Traffic Flow Forecasting}. In \bibinfo{booktitle}{\emph{Proceedings of the 29th ACM International Conference on Information \& Knowledge Management}} \emph{(\bibinfo{series}{CIKM '20})}. \bibinfo{publisher}{Association for Computing Machinery}, \bibinfo{address}{New York, NY, USA}, \bibinfo{pages}{1025--1034}.
\newblock
\showISBNx{978-1-4503-6859-9}
\urldef\tempurl%
\url{https://doi.org/10.1145/3340531.3411894}
\showDOI{\tempurl}


\bibitem[Lu et~al\mbox{.}(2016)]%
        {luIntegratingGrangerCausality2016}
\bibfield{author}{\bibinfo{person}{Zheng Lu}, \bibinfo{person}{Chen Zhou}, \bibinfo{person}{Jing Wu}, \bibinfo{person}{Hao Jiang}, {and} \bibinfo{person}{Songyue Cui}.} \bibinfo{year}{2016}\natexlab{}.
\newblock \showarticletitle{Integrating Granger Causality and Vector Auto-Regression for Traffic Prediction of Large-Scale WLANs}.
\newblock \bibinfo{journal}{\emph{KSII Transactions on Internet and Information Systems}} \bibinfo{volume}{10}, \bibinfo{number}{1} (\bibinfo{date}{Jan.} \bibinfo{year}{2016}), \bibinfo{pages}{136--151}.
\newblock


\bibitem[Luo et~al\mbox{.}(2021)]%
        {luo2021stan}
\bibfield{author}{\bibinfo{person}{Yingtao Luo}, \bibinfo{person}{Qiang Liu}, {and} \bibinfo{person}{Zhaocheng Liu}.} \bibinfo{year}{2021}\natexlab{}.
\newblock \showarticletitle{Stan: Spatio-temporal attention network for next location recommendation}. In \bibinfo{booktitle}{\emph{Proceedings of the Web Conference 2021}}. \bibinfo{pages}{2177--2185}.
\newblock


\bibitem[Lv et~al\mbox{.}(2015)]%
        {lvTrafficFlowPrediction2015}
\bibfield{author}{\bibinfo{person}{Yisheng Lv}, \bibinfo{person}{Yanjie Duan}, \bibinfo{person}{Wenwen Kang}, \bibinfo{person}{Zhengxi Li}, {and} \bibinfo{person}{Fei-Yue Wang}.} \bibinfo{year}{2015}\natexlab{}.
\newblock \showarticletitle{Traffic Flow Prediction With Big Data: A Deep Learning Approach}.
\newblock \bibinfo{journal}{\emph{IEEE Transactions on Intelligent Transportation Systems}} \bibinfo{volume}{16}, \bibinfo{number}{2} (\bibinfo{date}{April} \bibinfo{year}{2015}), \bibinfo{pages}{865--873}.
\newblock
\showISSN{1558-0016}
\urldef\tempurl%
\url{https://doi.org/10.1109/TITS.2014.2345663}
\showDOI{\tempurl}


\bibitem[Nie et~al\mbox{.}(2023)]%
        {nieTimeSeriesWorth2023}
\bibfield{author}{\bibinfo{person}{Yuqi Nie}, \bibinfo{person}{Nam~H Nguyen}, \bibinfo{person}{Phanwadee Sinthong}, {and} \bibinfo{person}{Jayant Kalagnanam}.} \bibinfo{year}{2023}\natexlab{}.
\newblock \showarticletitle{A Time Series is Worth 64 Words: Long-term Forecasting with Transformers}.
\newblock \bibinfo{journal}{\emph{arXiv preprint arXiv:2211.14730}} (\bibinfo{year}{2023}).
\newblock


\bibitem[Omar and Plapper(2020)]%
        {OmarInformationEntropy2020}
\bibfield{author}{\bibinfo{person}{Yamila~M. Omar} {and} \bibinfo{person}{Peter Plapper}.} \bibinfo{year}{2020}\natexlab{}.
\newblock \showarticletitle{A Survey of Information Entropy Metrics for Complex Networks}.
\newblock \bibinfo{journal}{\emph{Entropy}} \bibinfo{volume}{22}, \bibinfo{number}{12} (\bibinfo{date}{Dec.} \bibinfo{year}{2020}), \bibinfo{pages}{1417}.
\newblock
\showISSN{1099-4300}
\urldef\tempurl%
\url{https://doi.org/10.3390/e22121417}
\showDOI{\tempurl}


\bibitem[Park et~al\mbox{.}(2020)]%
        {ParkSTGRAT2020}
\bibfield{author}{\bibinfo{person}{Cheonbok Park}, \bibinfo{person}{Chunggi Lee}, \bibinfo{person}{Hyojin Bahng}, \bibinfo{person}{Yunwon Tae}, \bibinfo{person}{Seungmin Jin}, \bibinfo{person}{Kihwan Kim}, \bibinfo{person}{Sungahn Ko}, {and} \bibinfo{person}{Jaegul Choo}.} \bibinfo{year}{2020}\natexlab{}.
\newblock \showarticletitle{ST-GRAT: A Novel Spatio-Temporal Graph Attention Networks for Accurately Forecasting Dynamically Changing Road Speed}. In \bibinfo{booktitle}{\emph{Proceedings of the 29th ACM International Conference on Information \& Knowledge Management}} \emph{(\bibinfo{series}{CIKM '20})}. \bibinfo{publisher}{Association for Computing Machinery}, \bibinfo{address}{New York, NY, USA}, \bibinfo{pages}{1215--1224}.
\newblock
\showISBNx{978-1-4503-6859-9}
\urldef\tempurl%
\url{https://doi.org/10.1145/3340531.3411940}
\showDOI{\tempurl}


\bibitem[Rosvall and Bergstrom(2008)]%
        {RosvallMapsRandm2008}
\bibfield{author}{\bibinfo{person}{Martin Rosvall} {and} \bibinfo{person}{Carl~T Bergstrom}.} \bibinfo{year}{2008}\natexlab{}.
\newblock \showarticletitle{Maps of random walks on complex networks reveal community structure}.
\newblock \bibinfo{journal}{\emph{Proceedings of the national academy of sciences}} \bibinfo{volume}{105}, \bibinfo{number}{4} (\bibinfo{year}{2008}), \bibinfo{pages}{1118--1123}.
\newblock


\bibitem[Shang et~al\mbox{.}(2022)]%
        {ShangDiscreteGraphStructure2022}
\bibfield{author}{\bibinfo{person}{Chao Shang}, \bibinfo{person}{Jie Chen}, {and} \bibinfo{person}{Jinbo Bi}.} \bibinfo{year}{2022}\natexlab{}.
\newblock \showarticletitle{Discrete Graph Structure Learning for Forecasting Multiple Time Series}. In \bibinfo{booktitle}{\emph{International Conference on Learning Representations}}.
\newblock


\bibitem[Shannon(1948)]%
        {shannon1948mathematical}
\bibfield{author}{\bibinfo{person}{Claude~Elwood Shannon}.} \bibinfo{year}{1948}\natexlab{}.
\newblock \showarticletitle{A mathematical theory of communication}.
\newblock \bibinfo{journal}{\emph{The Bell system technical journal}} \bibinfo{volume}{27}, \bibinfo{number}{3} (\bibinfo{year}{1948}), \bibinfo{pages}{379--423}.
\newblock


\bibitem[Shao et~al\mbox{.}(2022)]%
        {shaoSpatialTemporalIdentitySimple2022}
\bibfield{author}{\bibinfo{person}{Zezhi Shao}, \bibinfo{person}{Zhao Zhang}, \bibinfo{person}{Fei Wang}, \bibinfo{person}{Wei Wei}, {and} \bibinfo{person}{Yongjun Xu}.} \bibinfo{year}{2022}\natexlab{}.
\newblock \showarticletitle{Spatial-Temporal Identity: A Simple yet Effective Baseline for Multivariate Time Series Forecasting}. In \bibinfo{booktitle}{\emph{Proceedings of the 31st ACM International Conference on Information \& Knowledge Management}} \emph{(\bibinfo{series}{CIKM '22})}. \bibinfo{publisher}{Association for Computing Machinery}, \bibinfo{address}{New York, NY, USA}, \bibinfo{pages}{4454--4458}.
\newblock
\showISBNx{978-1-4503-9236-5}
\urldef\tempurl%
\url{https://doi.org/10.1145/3511808.3557702}
\showDOI{\tempurl}


\bibitem[Sun et~al\mbox{.}(2020)]%
        {sun2020go}
\bibfield{author}{\bibinfo{person}{Ke Sun}, \bibinfo{person}{Tieyun Qian}, \bibinfo{person}{Tong Chen}, \bibinfo{person}{Yile Liang}, \bibinfo{person}{Quoc Viet~Hung Nguyen}, {and} \bibinfo{person}{Hongzhi Yin}.} \bibinfo{year}{2020}\natexlab{}.
\newblock \showarticletitle{Where to go next: Modeling long-and short-term user preferences for point-of-interest recommendation}. In \bibinfo{booktitle}{\emph{Proceedings of the AAAI Conference on Artificial Intelligence}}, Vol.~\bibinfo{volume}{34}. \bibinfo{pages}{214--221}.
\newblock


\bibitem[Vaswani et~al\mbox{.}(2017)]%
        {VaswaniAttention2017}
\bibfield{author}{\bibinfo{person}{Ashish Vaswani}, \bibinfo{person}{Noam Shazeer}, \bibinfo{person}{Niki Parmar}, \bibinfo{person}{Jakob Uszkoreit}, \bibinfo{person}{Llion Jones}, \bibinfo{person}{Aidan~N Gomez}, \bibinfo{person}{{\L}ukasz Kaiser}, {and} \bibinfo{person}{Illia Polosukhin}.} \bibinfo{year}{2017}\natexlab{}.
\newblock \showarticletitle{Attention Is All You Need}. In \bibinfo{booktitle}{\emph{Advances in Neural Information Processing Systems}}, Vol.~\bibinfo{volume}{30}. \bibinfo{publisher}{Curran Associates, Inc.}
\newblock


\bibitem[Wang et~al\mbox{.}(2021b)]%
        {wang2021gsnet}
\bibfield{author}{\bibinfo{person}{Beibei Wang}, \bibinfo{person}{Youfang Lin}, \bibinfo{person}{Shengnan Guo}, {and} \bibinfo{person}{Huaiyu Wan}.} \bibinfo{year}{2021}\natexlab{b}.
\newblock \showarticletitle{GSNet: learning spatial-temporal correlations from geographical and semantic aspects for traffic accident risk forecasting}. In \bibinfo{booktitle}{\emph{Proceedings of the AAAI conference on artificial intelligence}}, Vol.~\bibinfo{volume}{35}. \bibinfo{pages}{4402--4409}.
\newblock


\bibitem[Wang et~al\mbox{.}(2021a)]%
        {WangLibCity2021}
\bibfield{author}{\bibinfo{person}{Jingyuan Wang}, \bibinfo{person}{Jiawei Jiang}, \bibinfo{person}{Wenjun Jiang}, \bibinfo{person}{Chao Li}, {and} \bibinfo{person}{Wayne~Xin Zhao}.} \bibinfo{year}{2021}\natexlab{a}.
\newblock \showarticletitle{LibCity: An Open Library for Traffic Prediction}. In \bibinfo{booktitle}{\emph{Proceedings of the 29th International Conference on Advances in Geographic Information Systems}} \emph{(\bibinfo{series}{SIGSPATIAL '21})}. \bibinfo{publisher}{Association for Computing Machinery}, \bibinfo{address}{New York, NY, USA}, \bibinfo{pages}{145--148}.
\newblock
\showISBNx{978-1-4503-8664-7}
\urldef\tempurl%
\url{https://doi.org/10.1145/3474717.3483923}
\showDOI{\tempurl}


\bibitem[Wang et~al\mbox{.}(2020)]%
        {wang2020deep}
\bibfield{author}{\bibinfo{person}{Senzhang Wang}, \bibinfo{person}{Jiannong Cao}, {and} \bibinfo{person}{S~Yu Philip}.} \bibinfo{year}{2020}\natexlab{}.
\newblock \showarticletitle{Deep learning for spatio-temporal data mining: A survey}.
\newblock \bibinfo{journal}{\emph{IEEE transactions on knowledge and data engineering}} \bibinfo{volume}{34}, \bibinfo{number}{8} (\bibinfo{year}{2020}), \bibinfo{pages}{3681--3700}.
\newblock


\bibitem[Wang et~al\mbox{.}(2022)]%
        {wang2022knowledge}
\bibfield{author}{\bibinfo{person}{Yue Wang}, \bibinfo{person}{Mingsheng Liu}, \bibinfo{person}{Yongjian Huang}, \bibinfo{person}{Haifeng Zhou}, \bibinfo{person}{Xianhui Wang}, \bibinfo{person}{Senzhang Wang}, {and} \bibinfo{person}{Haohua Du}.} \bibinfo{year}{2022}\natexlab{}.
\newblock \showarticletitle{Knowledge-based and data-driven underground pressure forecasting based on graph structure learning}.
\newblock \bibinfo{journal}{\emph{International Journal of Machine Learning and Cybernetics}} (\bibinfo{year}{2022}), \bibinfo{pages}{1--16}.
\newblock


\bibitem[Wang et~al\mbox{.}(2019)]%
        {wang2019originmatrix}
\bibfield{author}{\bibinfo{person}{Yuandong Wang}, \bibinfo{person}{Hongzhi Yin}, \bibinfo{person}{Hongxu Chen}, \bibinfo{person}{Tianyu Wo}, \bibinfo{person}{Jie Xu}, {and} \bibinfo{person}{Kai Zheng}.} \bibinfo{year}{2019}\natexlab{}.
\newblock \showarticletitle{Origin-destination matrix prediction via graph convolution: a new perspective of passenger demand modeling}. In \bibinfo{booktitle}{\emph{Proceedings of the 25th ACM SIGKDD international conference on knowledge discovery \& data mining}}. \bibinfo{pages}{1227--1235}.
\newblock


\bibitem[Wu et~al\mbox{.}(2021)]%
        {wu2021autoformer}
\bibfield{author}{\bibinfo{person}{Haixu Wu}, \bibinfo{person}{Jiehui Xu}, \bibinfo{person}{Jianmin Wang}, {and} \bibinfo{person}{Mingsheng Long}.} \bibinfo{year}{2021}\natexlab{}.
\newblock \showarticletitle{Autoformer: Decomposition transformers with auto-correlation for long-term series forecasting}.
\newblock \bibinfo{journal}{\emph{Advances in Neural Information Processing Systems}}  \bibinfo{volume}{34} (\bibinfo{year}{2021}), \bibinfo{pages}{22419--22430}.
\newblock


\bibitem[Wu et~al\mbox{.}(2022)]%
        {wu2022structural}
\bibfield{author}{\bibinfo{person}{Junran Wu}, \bibinfo{person}{Xueyuan Chen}, \bibinfo{person}{Ke Xu}, {and} \bibinfo{person}{Shangzhe Li}.} \bibinfo{year}{2022}\natexlab{}.
\newblock \showarticletitle{Structural entropy guided graph hierarchical pooling}. In \bibinfo{booktitle}{\emph{Proceedings of the International Conference on Machine Learning}}. PMLR, \bibinfo{pages}{24017--24030}.
\newblock


\bibitem[Wu et~al\mbox{.}(2020)]%
        {wuConnectingDotsMultivariate2020}
\bibfield{author}{\bibinfo{person}{Zonghan Wu}, \bibinfo{person}{Shirui Pan}, \bibinfo{person}{Guodong Long}, \bibinfo{person}{Jing Jiang}, \bibinfo{person}{Xiaojun Chang}, {and} \bibinfo{person}{Chengqi Zhang}.} \bibinfo{year}{2020}\natexlab{}.
\newblock \showarticletitle{Connecting the {{Dots}}: {{Multivariate Time Series Forecasting}} with {{Graph Neural Networks}}}. In \bibinfo{booktitle}{\emph{Proceedings of the 26th {{ACM SIGKDD International Conference}} on {{Knowledge Discovery}} \& {{Data Mining}}}} \emph{(\bibinfo{series}{{{KDD}} '20})}. \bibinfo{publisher}{{Association for Computing Machinery}}, \bibinfo{address}{{New York, NY, USA}}, \bibinfo{pages}{753--763}.
\newblock
\showISBNx{978-1-4503-7998-4}
\urldef\tempurl%
\url{https://doi.org/10.1145/3394486.3403118}
\showDOI{\tempurl}


\bibitem[Wu et~al\mbox{.}(2019)]%
        {wuGraphWaveNetDeep2019}
\bibfield{author}{\bibinfo{person}{Zonghan Wu}, \bibinfo{person}{Shirui Pan}, \bibinfo{person}{Guodong Long}, \bibinfo{person}{Jing Jiang}, {and} \bibinfo{person}{Chengqi Zhang}.} \bibinfo{year}{2019}\natexlab{}.
\newblock \showarticletitle{Graph WaveNet for Deep Spatial-Temporal Graph Modeling}. In \bibinfo{booktitle}{\emph{Proceedings of the Twenty-Eighth International Joint Conference on Artificial Intelligence}}. \bibinfo{publisher}{Association for the Advancement of Artificial Intelligence (AAAI)}, \bibinfo{pages}{1907--1913}.
\newblock
\urldef\tempurl%
\url{https://doi.org/10.24963/ijcai.2019/264}
\showDOI{\tempurl}


\bibitem[Xia et~al\mbox{.}(2022)]%
        {xia2022multi}
\bibfield{author}{\bibinfo{person}{Jiangnan Xia}, \bibinfo{person}{Senzhang Wang}, \bibinfo{person}{Xiang Wang}, \bibinfo{person}{Min Xia}, \bibinfo{person}{Kun Xie}, {and} \bibinfo{person}{Jiannong Cao}.} \bibinfo{year}{2022}\natexlab{}.
\newblock \showarticletitle{Multi-view Bayesian spatio-temporal graph neural networks for reliable traffic flow prediction}.
\newblock \bibinfo{journal}{\emph{International Journal of Machine Learning and Cybernetics}} (\bibinfo{year}{2022}), \bibinfo{pages}{1--14}.
\newblock


\bibitem[Xu et~al\mbox{.}(2021)]%
        {xuSpatialTemporalTransformerNetworks2021}
\bibfield{author}{\bibinfo{person}{Mingxing Xu}, \bibinfo{person}{Wenrui Dai}, \bibinfo{person}{Chunmiao Liu}, \bibinfo{person}{Xing Gao}, \bibinfo{person}{Weiyao Lin}, \bibinfo{person}{Guo-Jun Qi}, {and} \bibinfo{person}{Hongkai Xiong}.} \bibinfo{year}{2021}\natexlab{}.
\newblock \bibinfo{booktitle}{\emph{Spatial-Temporal Transformer Networks for Traffic Flow Forecasting}}.
\newblock


\bibitem[Yang et~al\mbox{.}(2023)]%
        {yang2023minimum}
\bibfield{author}{\bibinfo{person}{Zhenyu Yang}, \bibinfo{person}{Ge Zhang}, \bibinfo{person}{Jia Wu}, \bibinfo{person}{Jian Yang}, \bibinfo{person}{Quan~Z Sheng}, \bibinfo{person}{Hao Peng}, \bibinfo{person}{Angsheng Li}, \bibinfo{person}{Shan Xue}, {and} \bibinfo{person}{Jianlin Su}.} \bibinfo{year}{2023}\natexlab{}.
\newblock \showarticletitle{Minimum entropy principle guided graph neural networks}. In \bibinfo{booktitle}{\emph{Proceedings of the Sixteenth ACM International Conference on Web Search and Data Mining}}. \bibinfo{pages}{114--122}.
\newblock


\bibitem[Yao et~al\mbox{.}(2019)]%
        {yaoRevisitingSpatialTemporalSimilarity2019}
\bibfield{author}{\bibinfo{person}{Huaxiu Yao}, \bibinfo{person}{Xianfeng Tang}, \bibinfo{person}{Hua Wei}, \bibinfo{person}{Guanjie Zheng}, {and} \bibinfo{person}{Zhenhui Li}.} \bibinfo{year}{2019}\natexlab{}.
\newblock \showarticletitle{Revisiting Spatial-Temporal Similarity: A Deep Learning Framework for Traffic Prediction}.
\newblock \bibinfo{journal}{\emph{Proceedings of the AAAI Conference on Artificial Intelligence}} \bibinfo{volume}{33}, \bibinfo{number}{01} (\bibinfo{date}{July} \bibinfo{year}{2019}), \bibinfo{pages}{5668--5675}.
\newblock
\showISSN{2374-3468}
\urldef\tempurl%
\url{https://doi.org/10.1609/aaai.v33i01.33015668}
\showDOI{\tempurl}


\bibitem[Ye et~al\mbox{.}(2021)]%
        {ye2021coupled}
\bibfield{author}{\bibinfo{person}{Junchen Ye}, \bibinfo{person}{Leilei Sun}, \bibinfo{person}{Bowen Du}, \bibinfo{person}{Yanjie Fu}, {and} \bibinfo{person}{Hui Xiong}.} \bibinfo{year}{2021}\natexlab{}.
\newblock \showarticletitle{Coupled layer-wise graph convolution for transportation demand prediction}. In \bibinfo{booktitle}{\emph{Proceedings of the AAAI conference on artificial intelligence}}, Vol.~\bibinfo{volume}{35}. \bibinfo{pages}{4617--4625}.
\newblock


\bibitem[Ying et~al\mbox{.}(2021)]%
        {ying2021transformers}
\bibfield{author}{\bibinfo{person}{Chengxuan Ying}, \bibinfo{person}{Tianle Cai}, \bibinfo{person}{Shengjie Luo}, \bibinfo{person}{Shuxin Zheng}, \bibinfo{person}{Guolin Ke}, \bibinfo{person}{Di He}, \bibinfo{person}{Yanming Shen}, {and} \bibinfo{person}{Tie-Yan Liu}.} \bibinfo{year}{2021}\natexlab{}.
\newblock \showarticletitle{Do transformers really perform badly for graph representation?}
\newblock \bibinfo{journal}{\emph{Advances in Neural Information Processing Systems}}  \bibinfo{volume}{34} (\bibinfo{year}{2021}), \bibinfo{pages}{28877--28888}.
\newblock


\bibitem[Yu et~al\mbox{.}(2018)]%
        {yuSpatioTemporalGraphConvolutional2018}
\bibfield{author}{\bibinfo{person}{Bing Yu}, \bibinfo{person}{Haoteng Yin}, {and} \bibinfo{person}{Zhanxing Zhu}.} \bibinfo{year}{2018}\natexlab{}.
\newblock \showarticletitle{Spatio-Temporal Graph Convolutional Networks: A Deep Learning Framework for Traffic Forecasting}. In \bibinfo{booktitle}{\emph{Proceedings of the Twenty-Seventh International Joint Conference on Artificial Intelligence}}. \bibinfo{publisher}{International Joint Conferences on Artificial Intelligence Organization}, \bibinfo{address}{Stockholm, Sweden}, \bibinfo{pages}{3634--3640}.
\newblock
\showISBNx{978-0-9992411-2-7}
\urldef\tempurl%
\url{https://doi.org/10.24963/ijcai.2018/505}
\showDOI{\tempurl}


\bibitem[Yuan et~al\mbox{.}(2018)]%
        {yuan2018hetero}
\bibfield{author}{\bibinfo{person}{Zhuoning Yuan}, \bibinfo{person}{Xun Zhou}, {and} \bibinfo{person}{Tianbao Yang}.} \bibinfo{year}{2018}\natexlab{}.
\newblock \showarticletitle{Hetero-convlstm: A deep learning approach to traffic accident prediction on heterogeneous spatio-temporal data}. In \bibinfo{booktitle}{\emph{Proceedings of the 24th ACM SIGKDD International Conference on Knowledge Discovery \& Data Mining}}. \bibinfo{pages}{984--992}.
\newblock


\bibitem[Zeng et~al\mbox{.}(2022)]%
        {zengAreTransformersEffective2022}
\bibfield{author}{\bibinfo{person}{Ailing Zeng}, \bibinfo{person}{Muxi Chen}, \bibinfo{person}{Lei Zhang}, {and} \bibinfo{person}{Qiang Xu}.} \bibinfo{year}{2022}\natexlab{}.
\newblock \bibinfo{booktitle}{\emph{Are Transformers Effective for Time Series Forecasting?}}
\newblock


\bibitem[Zeng et~al\mbox{.}(2023c)]%
        {zeng2023unsupervised}
\bibfield{author}{\bibinfo{person}{Guangjie Zeng}, \bibinfo{person}{Hao Peng}, \bibinfo{person}{Angsheng Li}, \bibinfo{person}{Zhiwei Liu}, \bibinfo{person}{Chunyang Liu}, \bibinfo{person}{Philip~S Yu}, {and} \bibinfo{person}{Lifang He}.} \bibinfo{year}{2023}\natexlab{c}.
\newblock \showarticletitle{Unsupervised Skin Lesion Segmentation via Structural Entropy Minimization on Multi-Scale Superpixel Graphs}.
\newblock \bibinfo{journal}{\emph{arXiv preprint arXiv:2309.01899}} (\bibinfo{year}{2023}).
\newblock


\bibitem[Zeng et~al\mbox{.}(2023a)]%
        {zeng2023effective}
\bibfield{author}{\bibinfo{person}{Xianghua Zeng}, \bibinfo{person}{Hao Peng}, {and} \bibinfo{person}{Angsheng Li}.} \bibinfo{year}{2023}\natexlab{a}.
\newblock \showarticletitle{Effective and Stable Role-Based Multi-Agent Collaboration by Structural Information Principles}. In \bibinfo{booktitle}{\emph{Proceedings of the AAAI conference on artificial intelligence}}, Vol.~\bibinfo{volume}{37}. \bibinfo{pages}{11772--11780}.
\newblock


\bibitem[Zeng et~al\mbox{.}(2023b)]%
        {zeng2023hierarchical}
\bibfield{author}{\bibinfo{person}{Xianghua Zeng}, \bibinfo{person}{Hao Peng}, \bibinfo{person}{Angsheng Li}, \bibinfo{person}{Chunyang Liu}, \bibinfo{person}{Lifang He}, {and} \bibinfo{person}{Philip~S Yu}.} \bibinfo{year}{2023}\natexlab{b}.
\newblock \showarticletitle{Hierarchical State Abstraction based on Structural Information Principles}. In \bibinfo{booktitle}{\emph{Proceedings of the 32nd International Joint Conference on Artificial Intelligence}}.
\newblock


\bibitem[Zhang et~al\mbox{.}(2020)]%
        {ZhangSpatioTemporalGraphStructure2020}
\bibfield{author}{\bibinfo{person}{Qi Zhang}, \bibinfo{person}{Jianlong Chang}, \bibinfo{person}{Gaofeng Meng}, \bibinfo{person}{Shiming Xiang}, {and} \bibinfo{person}{Chunhong Pan}.} \bibinfo{year}{2020}\natexlab{}.
\newblock \showarticletitle{Spatio-Temporal Graph Structure Learning for Traffic Forecasting}.
\newblock \bibinfo{journal}{\emph{Proceedings of the AAAI Conference on Artificial Intelligence}} \bibinfo{volume}{34}, \bibinfo{number}{01} (\bibinfo{date}{April} \bibinfo{year}{2020}), \bibinfo{pages}{1177--1185}.
\newblock
\showISSN{2374-3468}
\urldef\tempurl%
\url{https://doi.org/10.1609/aaai.v34i01.5470}
\showDOI{\tempurl}


\bibitem[Zhao et~al\mbox{.}(2020)]%
        {zhaoTGCNTemporalGraph2020}
\bibfield{author}{\bibinfo{person}{Ling Zhao}, \bibinfo{person}{Yujiao Song}, \bibinfo{person}{Chao Zhang}, \bibinfo{person}{Yu Liu}, \bibinfo{person}{Pu Wang}, \bibinfo{person}{Tao Lin}, \bibinfo{person}{Min Deng}, {and} \bibinfo{person}{Haifeng Li}.} \bibinfo{year}{2020}\natexlab{}.
\newblock \showarticletitle{T-GCN: A Temporal Graph Convolutional Network for Traffic Prediction}.
\newblock \bibinfo{journal}{\emph{IEEE Transactions on Intelligent Transportation Systems}} \bibinfo{volume}{21}, \bibinfo{number}{9} (\bibinfo{date}{Sept.} \bibinfo{year}{2020}), \bibinfo{pages}{3848--3858}.
\newblock
\showISSN{1558-0016}
\urldef\tempurl%
\url{https://doi.org/10.1109/TITS.2019.2935152}
\showDOI{\tempurl}


\bibitem[Zheng et~al\mbox{.}(2020)]%
        {zhengGMAN2020}
\bibfield{author}{\bibinfo{person}{Chuanpan Zheng}, \bibinfo{person}{Xiaoliang Fan}, \bibinfo{person}{Cheng Wang}, {and} \bibinfo{person}{Jianzhong Qi}.} \bibinfo{year}{2020}\natexlab{}.
\newblock \showarticletitle{GMAN: A Graph Multi-Attention Network for Traffic Prediction}. In \bibinfo{booktitle}{\emph{Proceedings of the AAAI Conference on Artificial Intelligence}} (2020-04-03), Vol.~\bibinfo{volume}{34}. \bibinfo{pages}{1234--1241}.
\newblock
Issue 01.
\showISSN{2374-3468}
\urldef\tempurl%
\url{https://doi.org/10.1609/aaai.v34i01.5477}
\showDOI{\tempurl}


\bibitem[Zhou et~al\mbox{.}(2021)]%
        {zhou2021informer}
\bibfield{author}{\bibinfo{person}{Haoyi Zhou}, \bibinfo{person}{Shanghang Zhang}, \bibinfo{person}{Jieqi Peng}, \bibinfo{person}{Shuai Zhang}, \bibinfo{person}{Jianxin Li}, \bibinfo{person}{Hui Xiong}, {and} \bibinfo{person}{Wancai Zhang}.} \bibinfo{year}{2021}\natexlab{}.
\newblock \showarticletitle{Informer: Beyond efficient transformer for long sequence time-series forecasting}. In \bibinfo{booktitle}{\emph{Proceedings of the AAAI conference on artificial intelligence}}, Vol.~\bibinfo{volume}{35}. \bibinfo{pages}{11106--11115}.
\newblock


\bibitem[Zou et~al\mbox{.}(2023)]%
        {zou2023se}
\bibfield{author}{\bibinfo{person}{Dongcheng Zou}, \bibinfo{person}{Hao Peng}, \bibinfo{person}{Xiang Huang}, \bibinfo{person}{Renyu Yang}, \bibinfo{person}{Jianxin Li}, \bibinfo{person}{Jia Wu}, \bibinfo{person}{Chunyang Liu}, {and} \bibinfo{person}{Philip~S Yu}.} \bibinfo{year}{2023}\natexlab{}.
\newblock \showarticletitle{SE-GSL: A General and Effective Graph Structure Learning Framework through Structural Entropy Optimization}. In \bibinfo{booktitle}{\emph{Proceedings of the ACM Web Conference 2023}}. \bibinfo{pages}{499--510}.
\newblock


\end{thebibliography}

\newpage
\appendix
\renewcommand\thefigure{\Alph{section}\arabic{figure}} 
\renewcommand\thetable{\Alph{section}\arabic{table}} 
%\renewcommand\theAlgoLine{}{\Alph{section}\arabic{AlgoLine}} 
% \renewcommand\thealgo{A.\arabic{Algorithm}}
% \SetAlgoCaptionSeparator{\Alph{section}}
\renewcommand\theequation{A.\arabic{equation}}
\setcounter{table}{0}
\setcounter{figure}{0}
\setcounter{equation}{0}

\end{document}